\documentclass[journal]{IEEEtranTIE}
\usepackage{graphicx}
\usepackage{cite}
\usepackage{picinpar}
\usepackage{amsmath}
\usepackage{url}
\usepackage{flushend}
\usepackage[latin1]{inputenc}
\usepackage{colortbl}
\usepackage{soul}
\usepackage{multirow}
\usepackage{pifont}
\usepackage{color}
\usepackage{alltt}
\usepackage[hidelinks]{hyperref}
\usepackage{enumerate}
\usepackage{siunitx}
\usepackage{breakurl}
\usepackage{epstopdf}
\usepackage{pbox}
\usepackage{url}
\usepackage{verbatim}
\usepackage{graphicx}
\usepackage{cite}
\usepackage{caption}
\usepackage{amssymb}
\usepackage{multirow}
\usepackage{adjustbox}
\usepackage{subfigure}
\usepackage{subcaption} 

\begin{document}
\title{TDANet: A Novel Temporal\\ Denoise Convolutional Neural\\ Network With Attention for Fault Diagnosis}

\author{
	\vskip 1em
	
	Zhongzhi Li, Rong Fan, Jingqi Tu, Jinyi Ma, Jianliang Ai and Yiqun Dong

	\thanks{
	
		Manuscript received Month xx, 2xxx; revised Month xx, xxxx; accepted Month x, xxxx.
		This work was sponsored by Shanghai Sailing Program (20YF1402500), Natural Science Foundation of Shanghai (22ZR1404500), and Shanghai Pujiang Program (22PJ1413800).
		
		Zhongzhi Li, Jingqi TU, Jinyi Ma, Jianliang Ai and Yiqun Dong are with the department of Aeronautics and Astronautics, Fudan University, Shanghai, 056004, China (e-mail: zzli22@m.fudan.edu.cn, jqtu23@m.fudan.edu.cn, jyma21@m.fudan.edu.cn, aijl@fudan.edu.cn and yiqundong@fudan.edu.cn). (corresponding author: Yiqun Dong).
		
		Rong Fan is with the school of Microelectronics, Shanghai University, Shanghai University, Shanghai, 201800, China (e-mail: rfan@shu.edu.cn).
	}
}

\maketitle

\begin{abstract}
Fault diagnosis plays a crucial role in maintaining the operational integrity of mechanical systems, preventing significant losses due to unexpected failures. As intelligent manufacturing and data-driven approaches evolve, Deep Learning (DL) has emerged as a pivotal technique in fault diagnosis research, recognized for its ability to autonomously extract complex features. However, the practical application of current fault diagnosis methods is challenged by the complexity of industrial environments. This paper proposed the Temporal Denoise Convolutional Neural Network With Attention (TDANet), designed to improve fault diagnosis performance in noise environments. This model transforms one-dimensional signals into two-dimensional tensors based on their periodic properties, employing multi-scale 2D convolution kernels to extract signal information both within and across periods. This method enables effective identification of signal characteristics that vary over multiple time scales. The TDANet incorporates a Temporal Variable Denoise (TVD) module with residual connections and a Multi-head Attention Fusion (MAF) module, enhancing the saliency of information within noisy data and maintaining effective fault diagnosis performance. Evaluation on two datasets, CWRU (single sensor) and Real aircraft sensor fault (multiple sensors), demonstrates that the TDANet model significantly outperforms existing deep learning approaches in terms of diagnostic accuracy under noisy environments.
\end{abstract}


\markboth{IEEE TRANSACTIONS ON INDUSTRIAL ELECTRONICS}%
{}

\definecolor{limegreen}{rgb}{0.2, 0.8, 0.2}
\definecolor{forestgreen}{rgb}{0.13, 0.55, 0.13}
\definecolor{greenhtml}{rgb}{0.0, 0.5, 0.0}

\section{Introduction}
\IEEEPARstart{M}{echanical} systems (typically such as rolling bearings, aircraft sensors, etc.) are widely used in intelligent devices \cite{yang2019polynomial}. Since mechanical systems often work in harsh conditions, including high temperature, high humidity, and overload, which are prone to failure \cite{he2021modified}. These failures can cause economic losses and safety risks, and even lead to casualties \cite{chen2023novel}. Therefore, accurate monitoring and diagnosis of mechanical system faults are crucial for maintaining production safety and preventing catastrophic incidents \cite{chen2021gaussian}.

Traditional fault diagnosis are usually based on system models or signal processing models. Jalan et al. proposed a model-based fault diagnosis method for rolling bearing systems \cite{jalan2009model}. Dyba{\l}a et al. proposed a signal processing-based diagnosis method. The pure noise part is extracted from the original vibration signal through Empirical Mode Decomposition (EMD), and the spectral analysis method of the empirically determined local amplitude is used to further extract faults relevant features from the pure noise signal \cite{dybala2014rolling}. These methods largely depend on manually identifying intricate features, heavily rely on empirical knowledge, which leads to reduced efficiency and narrow applicability of the algorithms.

Machine learning methods can automatically extract data features and have been widely applied in fault diagnosis tasks \cite{zhang2020deep}. Typical methods include Artificial Neural Network (ANN), Principal Component Analysis (PCA), K-Nearest Neighbors (KNN) and Support Vector Machine (SVM), etc. Zarei et al. proposed an ANN system method for processing the time domain characteristics of fault signals \cite{zarei2012induction}. Qian et al. used a differential evolution algorithm to preprocess time domain signals and improved the feature extraction method \cite{qian2016application}. Xia et al. used PCA to reduce the dimensionality of the preprocessed signal, and obtained fault diagnosis results through the optimal classifier sequence selected by the decision fusion algorithm \cite{6030215}. Xue et al. proposed an improved Local and Global Principle Component Analysis (LGPCA) method, and used the fast Ensemble Empirical Mode Decomposition (EEMD) for multiscale feature extraction in fault diagnosis \cite{xue2017hybrid}. The fault diagnosis methods based on machine learning  avoid the limitations of manually identifying intricate features. However, the methods based on machine learning are often difficult to extract high-dimensional features from data due to its simple structure (fewer network layers, etc.), which limits the diagnostic performance of the algorithm.

Deep learning algorithms have more complex network structures and can learn richer feature representations from data, which have been widely used in fault diagnosis tasks \cite{miao2021sparse}. Janssens et al. proposed an equipment condition monitoring model based on Convolutional Neural Network (CNN), which automatically extracts data features \cite{janssens2016convolutional}. Xu et al. proposed a hybrid deep learning model based on CNN and deep forest, which processed signals through Continuous Wavelet Transformation (CWT) \cite{xu2021hybrid}. Mao et al. proposed a new type of Deep AutoEncoder model (DAE) that combines various types of information, has effective land acquisition capabilities, and has simplified the DAE network structure \cite{mao2021new}. Ma et al. proposed a multi-objective optimization integration algorithm \cite{ma2019ensemble}, which weighted the integration of Convolutional Residual Network (CRN), Deep Belief Network (DBN) and DAE, showing better adaptability compared to single models. However, the above studies mainly focus on the design and optimization of diagnostic algorithms, overlooking the significant noise included in the collected data in practical scenarios, which has a serious impact on the performance of diagnostic algorithms.

In recent years, deep learning methods for solving fault diagnosis in noisy environments have been widely studied. Chen et al. used two CNNs with different kernel sizes to automatically extract signal features of different frequencies from the original data, and then used Long Short-Term Memory (LSTM) to identify the fault type \cite{chen2021bearing}. Compared with other intelligent algorithms, it has better diagnosis performance between the Signal-to-Noise Ratio (SNR) of -2 dB and 10 dB. Zhang et al. used residual learning to improve network training and conducted experiments with SNR between 0 dB and 8 dB \cite{zhang2019deep}. Liang et al. combined wavelet transform with improved Resnet and proposed a new rolling bearing fault diagnosis method, which is robust to noise \cite{liang2022intelligent}. Guo et al proposed a method based on Attention CNN and BiLSTM (ACNN-BiLSTM) \cite{guo2023rolling}. Short-term spatial features are extracted through Attention CNN, and BiLSTM is used to extract long-distance dependence information of signals. The setting range of the SNR in the experiment is 10 dB to 20 dB.

To further improve the diagnostic performance of deep learning models in noise environments (especially when SNR $\textless$ 0), this paper proposes a Novel Temporal Denoise Convolutional Neural Network (TDANet). The main highlights of this paper are as follows:

(1) Efficient multi-time-resolution signal feature extraction: In order to fully extract features data mixed with noise, the Short-Time Fourier Transform (STFT) is used to decompose the signal to obtain different frequencies and amplitudes, which is suitable for processing non-stationary signals. Different frequencies of signal sorted by amplitude values are stacked into 2D images respectively, and multi-scale 2D convolution is used to extract features on the 2D image (including intra-period and inter-period features). This is different from the signal-image mapping method proposed by Zhao J et al. \cite{zhao2021new}, which is a single-time-resolution method and lack in considering the changing characteristics of different signals and signals at different times.

(2) Nonlinear noise filtering: The threshold unit in the Temporal Variable Denoise (TVD) module can dynamically and flexibly apply nonlinear processing to the data signal, reducing noise in the signal. This dynamic adjustment ensures optimal noise reduction across a broad spectrum of signal types and noise levels, without the need for manual intervention or preset thresholds that may not be suitable for all conditions.

(3) Multi-head attention fusion weighting: The multi-head attention mechanism is used to dynamically weight the components of the signal at different frequencies. Different from using amplitude values at different frequencies as weighting weights, each head of the multi-head attention mechanism can focus on different features of the signal, which can learn rich and diverse feature representations.

(4) Multi-scenario experimental verification: This paper conducts a series of comparative experiments and ablation experiments using the CWRU rolling bearing dataset (single-sensor signal) and Real aircraft sensor dataset (multiple sensors). The experimental results show that the proposed TDANet can maintain satisfactory classification performance when performing fault diagnosis tasks in noise environments, outperforming other compared deep learning models.

The rest part of this paper is organized as follows. Section II details the principles of the proposed TDANet and the sub-modules. Section III conducts experimental verification. Finally, conclusions are drawn in Section IV.

\section{Methodology and proposed framework}

\begin{figure*}[htbp]
\centering
\includegraphics[clip,width = 0.90
\linewidth]{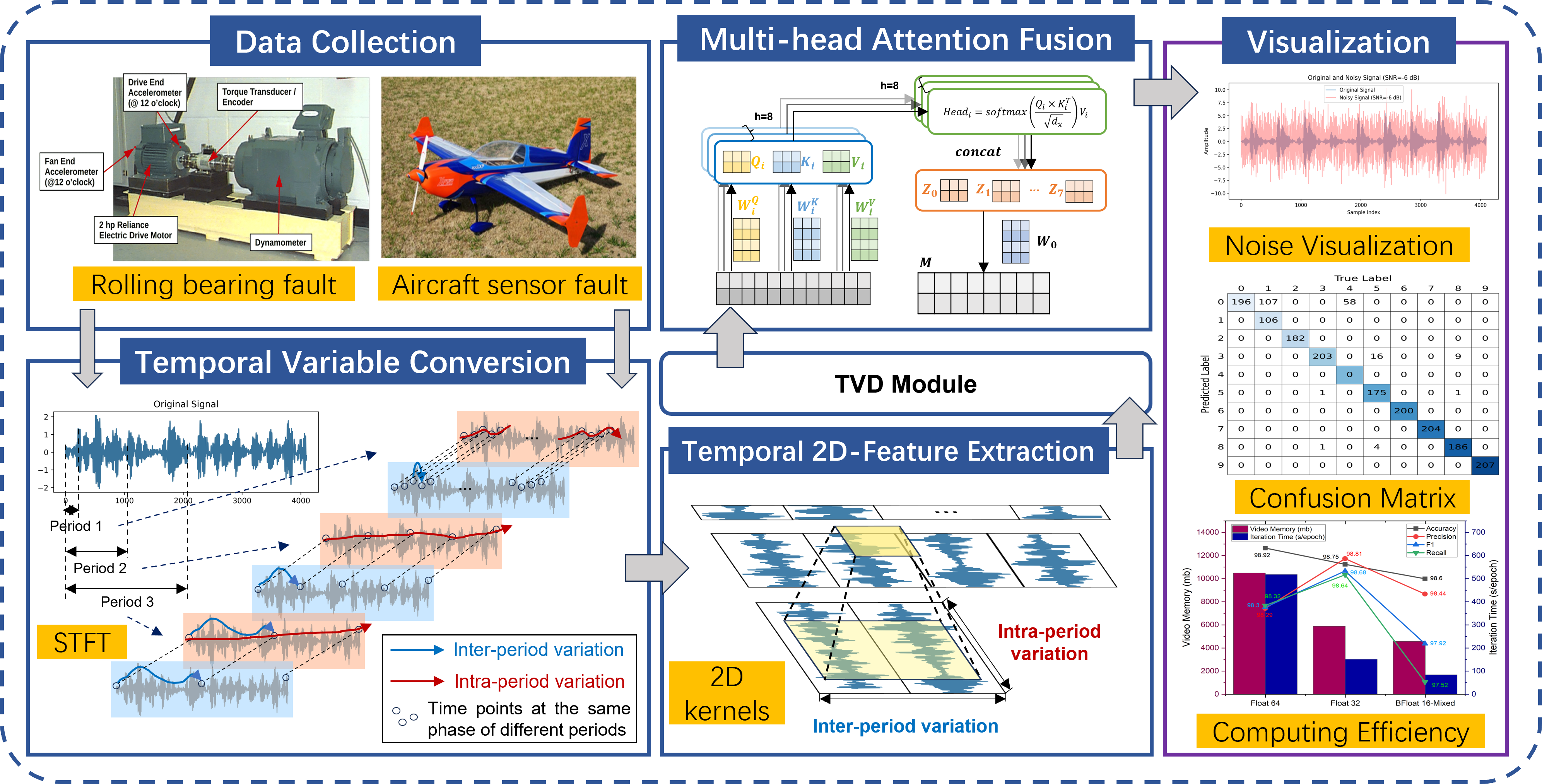}
\caption{Framework of the proposed TDANet model.}
\label{Fig:Framework}
\end{figure*}

The TDANet framework proposed in this paper is shown in Fig. \ref{Fig:Framework}. Its core modules include Temporal Variable Conversion (TVC), Temporal 2D-Feature Extraction (TFE), Temporal Variable Denoise (TVD), and Multi-head Attention Fusion (MAF), aiming to reduce the noise in the signal and improve the classification performance of the faults. The individual modules of the framework will be discussed in detail in Sections \ref{TVCAND2D}-\ref{TSP}.

\subsection{Temporal Variable Conversion and 2D-Feature Extraction Modules}
\label{TVCAND2D}
\begin{figure}[htbp]
\centering
\includegraphics[clip,width = 1.0
\linewidth]{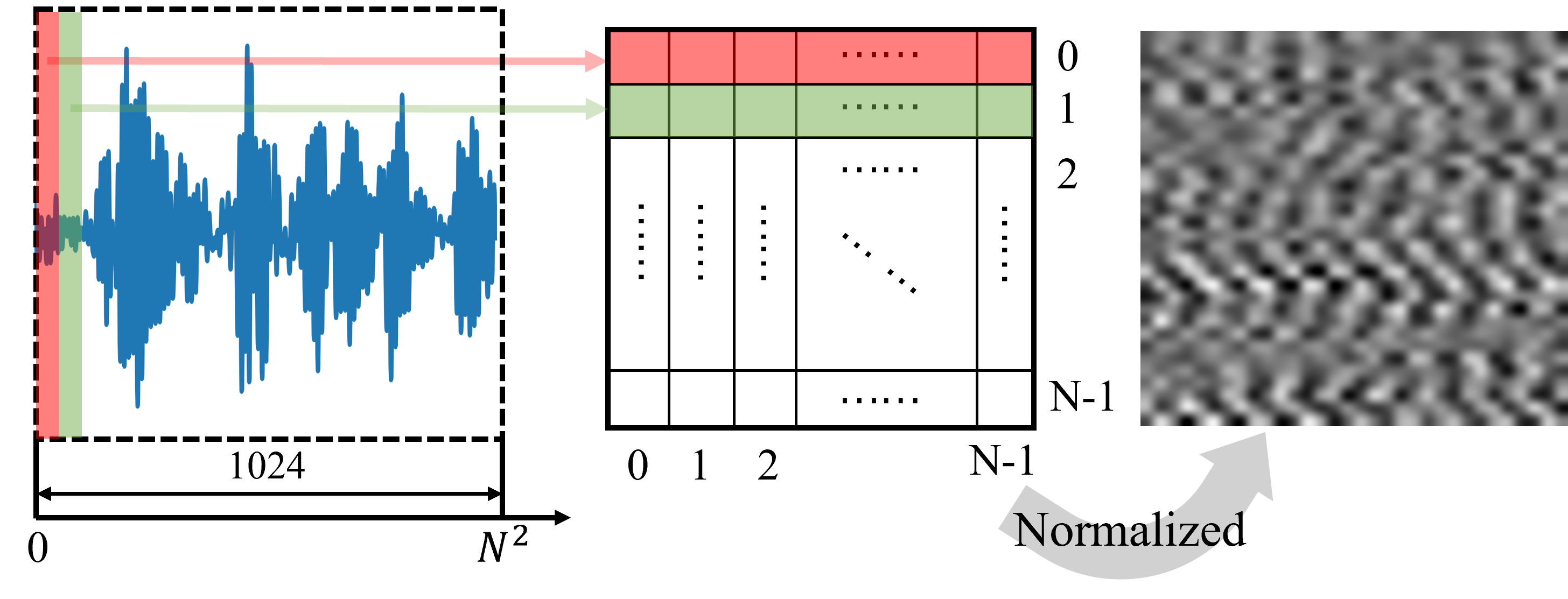}
\caption{The traditional signal stacking method.}
\label{Fig:Old_ToImage}
\end{figure}

It is difficult to extract features directly from one-dimensional signals with noise. A common method is to stack one-dimensional signals into two-dimensional images, which is beneficial for algorithms to extract coupling information from signals. The traditional signal stacking method is shown in Fig. \ref{Fig:Old_ToImage}. Assuming the number of sampling points of the one-dimensional signal is $N^{2}$, the principle involves decomposing the signal into $N$ segments, each with $N$ sampling points, and then stacking these $N$ segments into an $N \times N$ two-dimensional tensor. This stacking method can be called single-time-resolution stacking, where the value of $N$ is fixed and does not consider distinctions between different signals.

\begin{figure}[htbp]
\centering
\includegraphics[clip,width = 1.0
\linewidth]{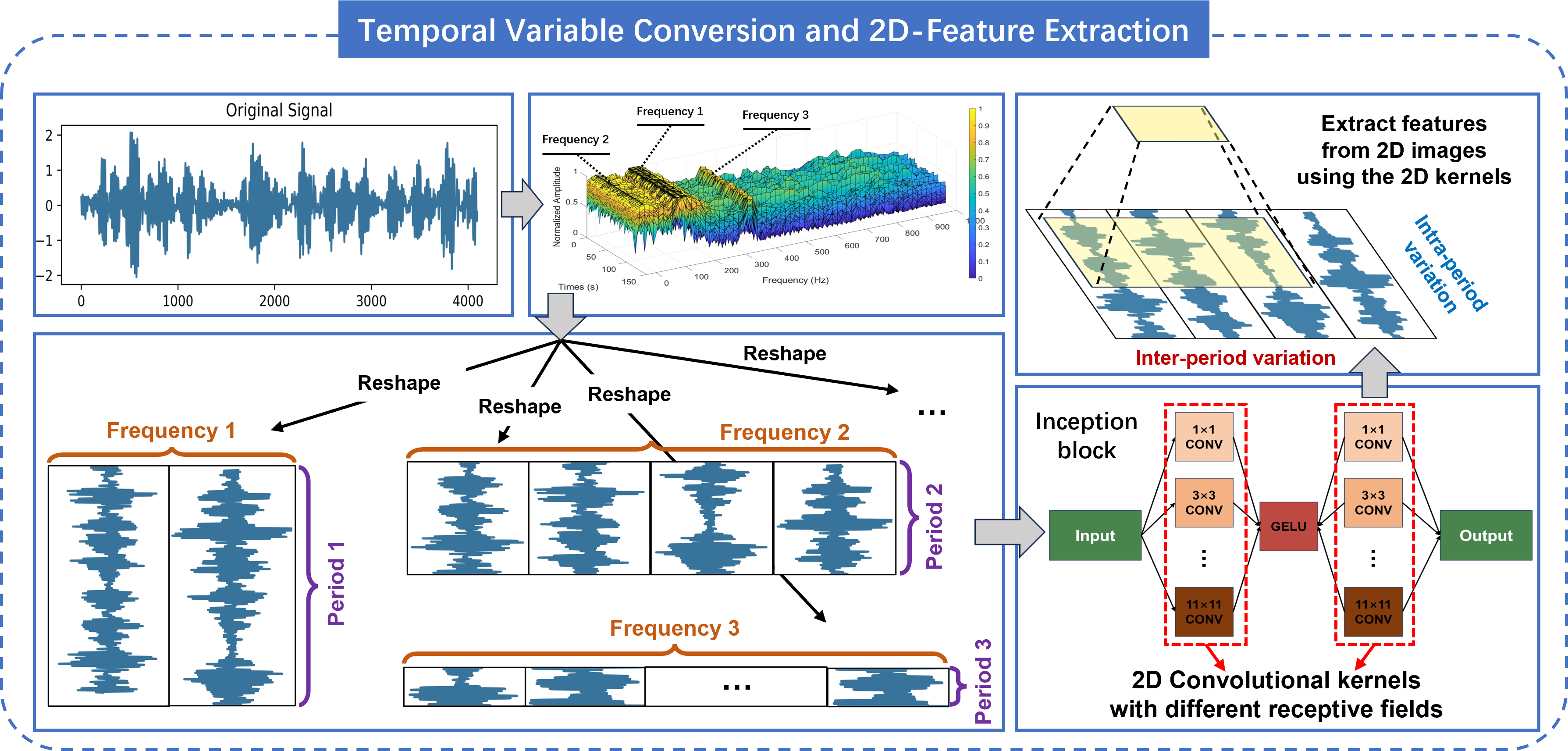}
\caption{Signal stacking with multiple temporal resolutions and multi-scale feature extraction.}
\label{Fig:TVC}
\end{figure}

In order to improve the signal stacking method, this paper proposed an multi-time-resolution signal stacking method, which uses STFT to decompose the one-dimensional signal into multiple periods, and stacks the signals of different periods respectively, as shown in Fig. \ref{Fig:TVC}.

STFT is a time-frequency domain analysis method for time-varying signals \cite{sandsten2016time}. Compared with Fast Fourier Transform (FFT), STFT is better suited for handling non-stationary signals. Essentially, a time-limited window function, denoted as $h(t)$, is introduced to the signal before undergoing Fourier transformation in STFT. For a signal with length $T$ and sampling points $C$, its original sequence is represented as $X_{1D} \in \mathbb{R}^{T \times C}$. In this paper, the original one-dimensional signal $X_{1D}$ is assumed to be constant within a certain short period of time, and the window function $h(t)$ moves on the signal to convert it by segment. The calculation formula of STFT is given by equation (\ref{equation_1}).
\begin{equation}
STFT\left( {f(t)} \right) = {\int_{- \infty}^{+ \infty}{f(t)h\left( {t - \tau} \right)e^{- j\omega t}dt}}
\label{equation_1}
\end{equation}
where $STFT(*)$ represents calculation of the STFT; $f(t)$ is the time domain signal before transformation, $h\left( {t - \tau} \right)$ is the window function, and $\tau$ is the center of the window function. The calculation equations for using STFT to analyze time series in the frequency domain is shown in equations (\ref{equation_2}-\ref{equation_4}).
\begin{equation}
\begin{aligned}
\mathbf{A} &= Avg\left(Amp\left(STFT\left( X_{1D} \right) \right) \right)
\end{aligned}
\label{equation_2}
\end{equation}

\begin{equation}
\begin{aligned}
\quad \left\{ f_{1}, \ldots, f_{k} \right\} =
argTopk_{f_{*} \in \left\{ 1, \ldots, \left\lfloor \frac{T}{2} \right\rfloor \right\}}\left( \mathbf{A} \right)
\end{aligned}
\label{equation_3}
\end{equation}

\begin{equation}
\begin{aligned}
p_{i} = \left\lfloor \frac{T}{f_{i}} \right\rfloor, \quad i \in \left\{ 1, \ldots, k \right\}
\end{aligned}
\label{equation_4}
\end{equation}
where $Amp(*)$ represents the calculation of amplitude value. $Avg(*)$ indicates that the calculated amplitude of a frequency is averaged over $C$ dimensions. Due to the conjugation of the frequency domain, this paper only considers frequencies within the range $\left\{ {1,~\ldots,~\left\lbrack \frac{T}{2} \right\rbrack} \right\}$. To avoid introducing noise from high-frequency signals, this paper only selects the top $k$ highest amplitude values, and obtains the most significant $k$ frequencies through non-normalized amplitudes $~\left\{ {\mathbf{A}_{f1},~\ldots,~\mathbf{A}_{fk}} \right\}$ (where $\mathbf{A}_{f_n}$ represents the intensity of the periodic basis function at frequency $f_n$), along with their corresponding $k$ periods, where $k$ is a hyperparameter. Therefore, equations (\ref{equation_2}-\ref{equation_4}) can be reformulated as
\begin{equation}
Period\left( \mathbf{X}_{1D} \right) = ~\mathbf{A},~\left\{ {f_{1},~\ldots,~f_{k}} \right\},~\left\{ {p_{1},~\ldots,~p_{k}} \right\}
\label{equation_5}
\end{equation}

Based on the selected frequencies $\left\{ {f_{1},~\ldots,~f_{k}} \right\}$ and periods $\left\{ {p_{1},~\ldots,~p_{k}} \right\}$, this paper utilizes the equation (\ref{equation_6}) to reshape the one-dimensional time series $\mathbf{X}_{1D}$ into two-dimensional tensors.
\begin{equation}
X_{2D}^{i} = {Reshape}_{p_{i},~f_{i}}\left( {sequence\left( \mathbf{X}_{1D} \right)} \right)
\label{equation_6}
\end{equation}
where $sequence(*)$ is to unfold the time series to adaptively fills the $Reshape$; $p_i$ and $f_i$ respectively represent the number of rows and columns of the two-dimensional tensor; $X_{2D}^{i} \in \mathbb{R}^{p_{i} \times f_{i} \times C}$ denotes the two-dimensional tensor of the frequency $f_i$ corresponding to the one-dimensional sequence, recorded as the $i$-th tensor. With multiple different $p_i$ and $f_i$, a sequence of two-dimensional tensors $\left\{ {X_{2D}^{1},~\ldots,~X_{2D}^{k}} \right\}$ is obtained. As shown in Fig. \ref{Fig:TVC}, the columns and rows indicate the intraperiod-variation and interperiod-variation of the signal at the corresponding frequencies.

After obtaining the sequence of two-dimensional tensors, this paper performs feature extraction by constructing an Inception block containing multi-scale 2D convolutional kernels. The calculation formula is shown in Equation (\ref{equation_7}).
\begin{equation}
{\hat{X}}_{2D}^{l,~i} = Inception\left( X_{2D}^{i} \right)
\label{equation_7}
\end{equation}

\subsection{Temporal Variable Denoise Module}
\label{TVD}
\begin{figure}[htbp]
\centering
\includegraphics[clip,width = 0.8
\linewidth]{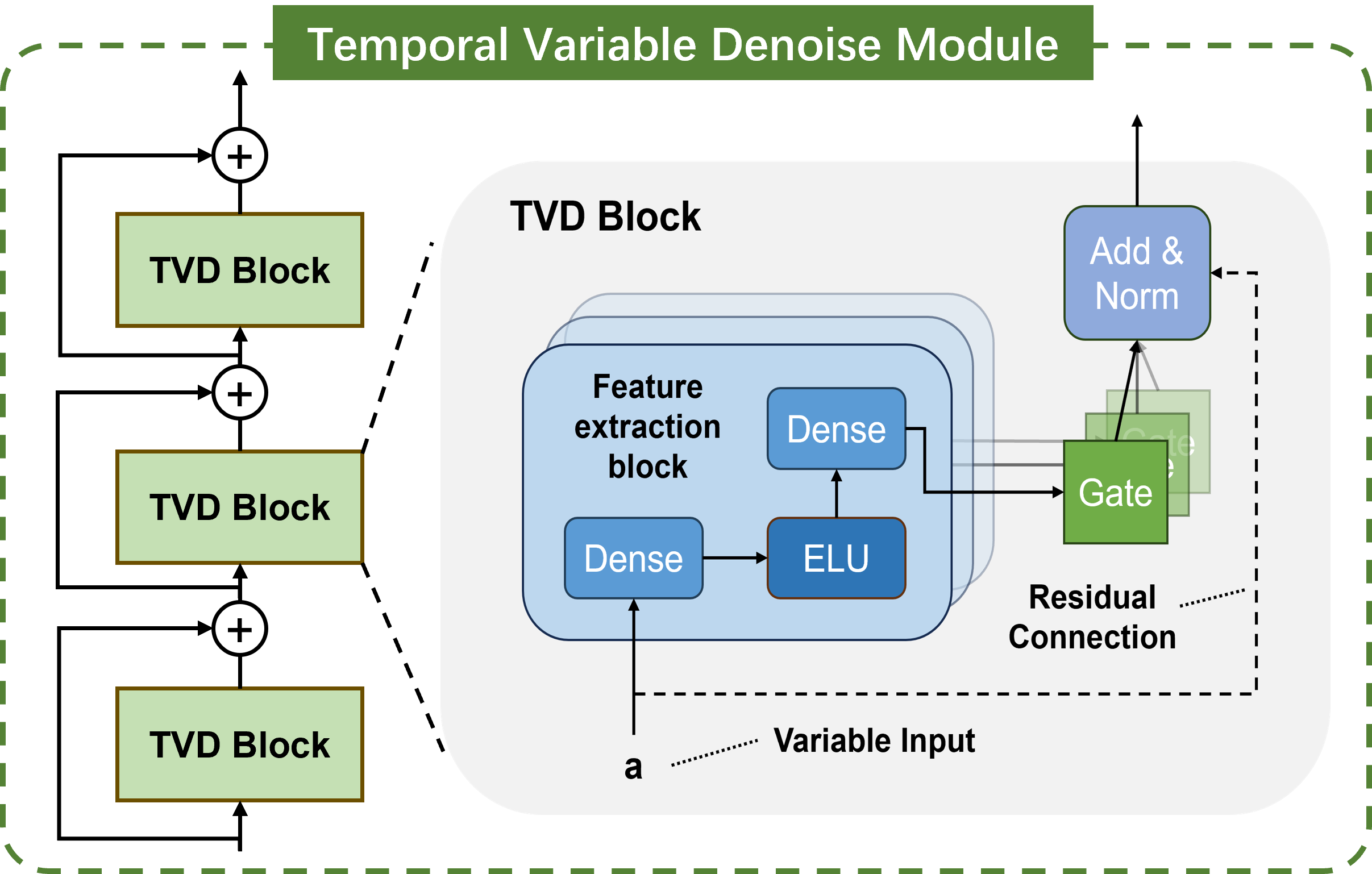}
\caption{TVD module that can perform non-linear signal filtering.}
\label{Fig:TVD_Module}
\end{figure}

In order to dynamically and flexibly reduce the noise in the input signal, the TVD module is designed in this paper, as shown in Fig. \ref{Fig:TVD_Module}. For the $l$-th TVD Block, the computation equation is given by

\begin{equation}
{TVD}_{\omega}\left( {\hat{X}}_{2D}^{l,~i} \right) = LayerNorm\left( {{\hat{X}}_{2D}^{l,~i} + {GLU}_{\omega}\left( \eta_{1} \right)} \right)
\label{equation_8}
\end{equation}

\begin{equation}
\eta_{1} = W_{1,~\omega}\eta_{2} + b_{1,~\omega}
\label{equation_9}
\end{equation}

\begin{equation}
\eta_{2} = ELU\left( {W_{2,~\omega}{\hat{X}}_{2D}^{l,~i} + b_{2,~\omega}} \right)
\label{equation_10}
\end{equation}
where $\eta_{1} \in \mathbb{R}^{d_{model}}$ and $\eta_{2} \in \mathbb{R}^{d_{model}}$ are hidden layers, and $d_{model}$ is the dimension of the hidden layer. $W_{*} \in \mathbb{R}^{d_{model} \times d_{model}}$ is the weights, $b_{*} \in \mathbb{R}^{d_{model}}$ is the bias, and $\omega$ represents the shared parameters of the weights. $LayerNorm(*)$ is standard layer normalization. $ELU(*)$ is the Exponential Linear Unit activation function. When $W_{2,~\omega}{\hat{X}}_{2D}^{l,~i} + b_{2,~\omega} \gg 0$, the ELU activation function will act as an identity function; when $W_{2,~\omega}{\hat{X}}_{2D}^{l,~i} + b_{2,~\omega} \ll 0$, the ELU activation function will produce a constant output. ${GLU}_{\omega}(*)$ is a gating layer based on gated linear units, which function is to suppress the unnecessary parts (noise in this paper). When $\beta \in \mathbb{R}^{d_{model}}$ is taken as input, the expression of GLU is as
\begin{equation}
{GLU}_{\omega}(\beta) = \sigma\left( {W_{3,~\omega}\beta + b_{3,~\omega}} \right)\bigodot\left( {W_{4,~\omega}\beta + b_{4,~\omega}} \right)
\label{equation_11}
\end{equation}
where $\sigma(*)$ is the sigmoid activation function, and $\bigodot$ is the Hadamard product of elements. GLU can be used to control the passage of data information flow in TVD blocks, thereby achieving the purpose of nonlinear filtering of signal.

\subsection{Multi-head Attention Fusion Module}
\label{MAF}
$\hat{X}_{1D}^{L,~k} = \left\{ {{\hat{X}}_{1D}^{l,~1},~\ldots,~{\hat{X}}_{1D}^{l,~k}} \right\}$ is the output vector after denoising of the TVD module ($L$ is the total number of TVD blocks). This paper adopts the MAF module to dynamically weight the components of the signal at different frequencies. Different from using the amplitudes in the frequency domain as the weight, the proposed weighting method can effectively capture the critical information.

$Q_{i}$ (Query), $K_{i}$ (Key), and $V_{i}$ (Value) matrices are used to calculate the attention weights respectively, where $Q_{i}=\hat{X}_{1D}^{L,~k} W_{i}^{Q}$, $K_{i}=\hat{X}_{1D}^{L,~k} W_{i}^{K}$ and $V_{i}=\hat{X}_{1D}^{L,~k} W_{i}^{V}$ ($i$ represents the $i$-th head). The calculation equation of the attention weights of the $i$-th head is as follows
\begin{equation}
{Attention}_{i}\left( {Q_{i},~K_{i},~V_{i}} \right) = softmax\left( \frac{Q_{i}K_{i}^{T}}{d_{attn}} \right)V_{i}
\label{equation_12}
\end{equation}
where $d_{attn}$ is used to scale the inner product to avoid the input of the $softmax(*)$ function being too large or too small. Concatenate the output vectors of multiple heads to obtain the matrix $Z$.
\begin{equation}
Z = concat\left( {{Attention}_{1},~\ldots,~{Attention}_{i}} \right)
\label{equation_13}
\end{equation}

A linear transformation is applied to the matrix $Z$ to obtain the weighted feature information output $h_{attn}$.
\begin{equation}
h_{attn} = Z \times W
\label{equation_14}
\end{equation}
where $W$ is the weight matrix that can be trained.

\subsection{Training Strategy of the Proposed Model}
\label{TSP}
\begin{figure}[htbp]
\centering
\includegraphics[clip,width = 1.0
\linewidth]{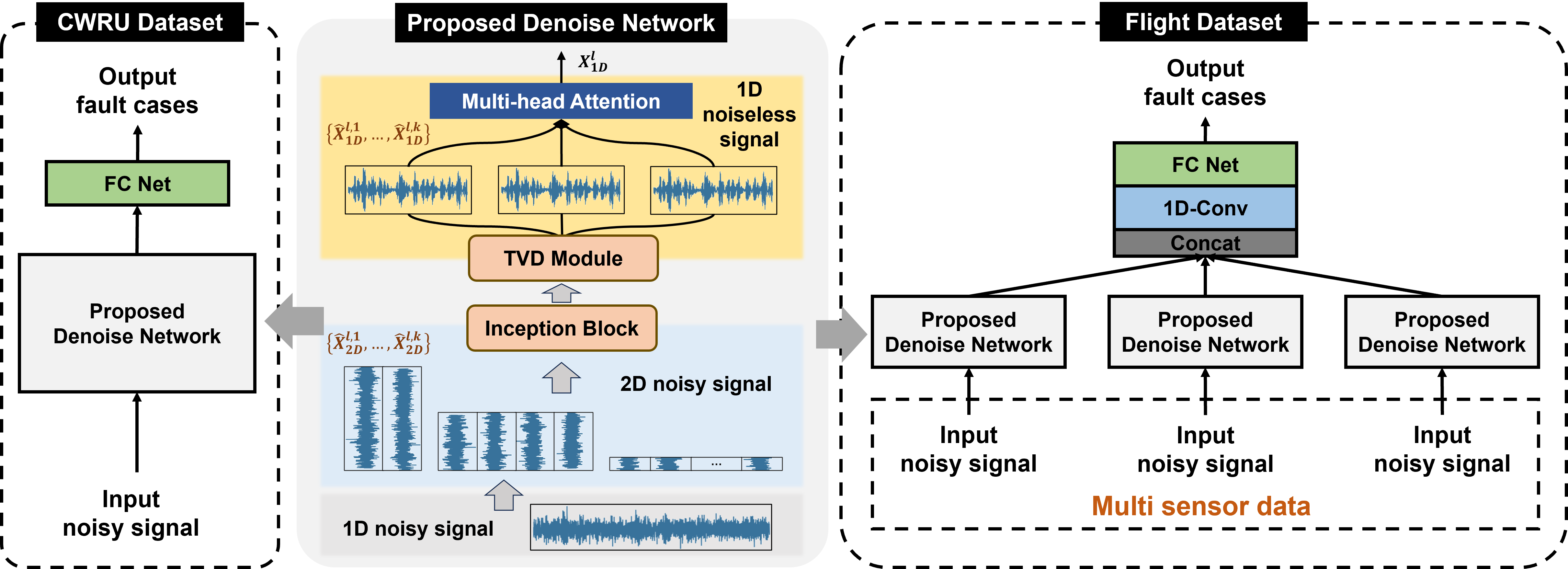}
\caption{The forward propagation process of TDANet.}
\label{Fig:PDNforDAF}
\end{figure}

The forward propagation process of the proposed TDANet for both single-sensor and multi-sensor data during training is illustrated in Fig. \ref{Fig:PDNforDAF}. For the rolling bearing data collected by a single sensor, the 2D images of the signal at different time resolutions are obtained after STFT decomposition. Then the Inception Block with multi-scale 2D convolution kernel can be used to captured the feature information. The TVD module provides a channel for denoising the feature information. The features are weighted by the MAF module to obtain a one-dimensional tensor, which is then processed through the linear layer of the neural network to obtain the category of the fault. For real aircraft attitude measurement data with multiple sensors, the TDANet is utilized to perform denoising on different sensors. Finally, the multiple denoised one-dimensional tensors are spliced as the input of the neural network classification layer, yielding the category of the fault.

\begin{figure}[htbp]
\centering
\includegraphics[clip,width = 0.95
\linewidth]{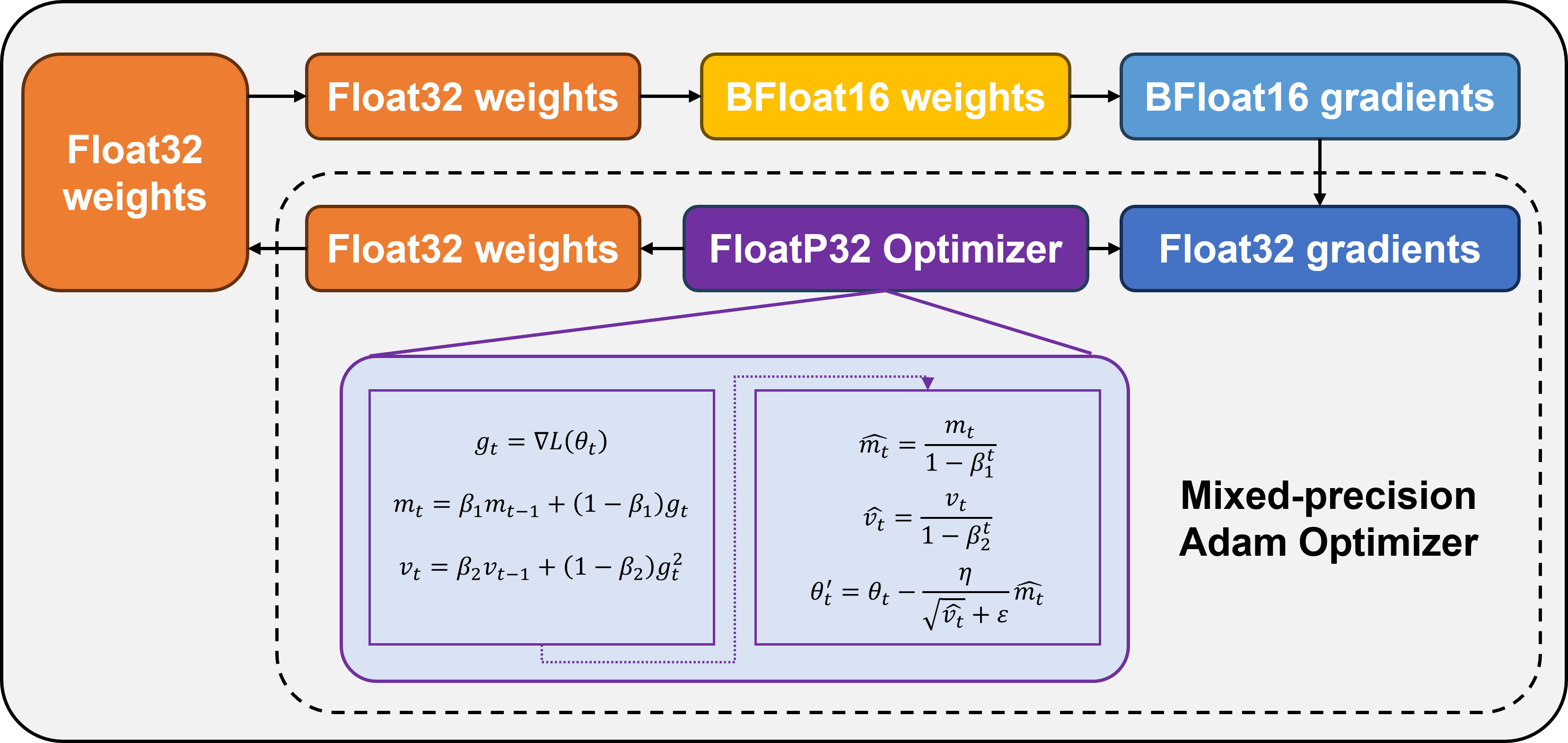}
\caption{Mixed precision calculation method used by TDANet.}
\label{Fig:Training_Step}
\end{figure}

To reduce computational costs, increase training speed, and save computing resource space while maintaining model performance, this paper chooses the optimization strategy of mixed precision calculation in the TDANet model training, as depicted in Fig. \ref{Fig:Training_Step}. During the forward propagation of the model training, BFloat16 precision parameters are utilized. During the backward propagation of gradients, to avoid numerical underflow caused by gradient scaling, this paper employs Adam optimizer with Float32 precision.

\section{Experimental verification}

In this section, the performance of the proposed model is verified on the dataset of Case Western Reserve University Bearing Data \cite{hendriks2022towards} and the real aircraft sensor fault \cite{li2023lightweight}. The code is written on the platform of Data Science Workshop with Python 3.8. The training of the model is completed on an integrated development platform, which has one Nvidia-A100 GPU with 80 GB video memory.

\subsection{Evaluation Metrics}
Following common settings, 4 evaluation metrics (Accuracy, Precision, F1 and Recall) are used to evaluate the performance of the model \cite{liu2021machinery}. Table \ref{table:Confusion_matrix} depicts the correspondence between the predictions of TDANet and the true labels.
\begin{table}[htbp]
	\centering
	\caption{Confusion matrix of predictions and labels.}
	\label{table:Confusion_matrix}
	\begin{tabular}{ccccc}
		\hline
		\hline
        &Predictions (Positive)       &Predictions (Negative)       \\
        \hline
        Labels (Positive)	        & TP      & FN       \\
        Labels (Negative)	        & FP   & TN       \\
	\hline
    \hline
	\end{tabular}
\end{table}

Expanding to the multi-classification tasks (taking $n$ classes as an example), the values of $\overrightarrow{TP}$ (True Positive), $\overrightarrow{FP}$ (False Positive), $\overrightarrow{FN}$ (False Negative), and $\overrightarrow{TN}$ (True Negative) are $n$-dimensional vectors, where $n$ represents the number of classes in the dataset (in this study, $n$ = 10 and 6, respectively). The calculation formulas for evaluation metrics are as follows:
\begin{equation}
Accuracy = \left| \frac{\overrightarrow{TP}+\overrightarrow{TN}}{\overrightarrow{TP}+\overrightarrow{TN}+\overrightarrow{FN}+\overrightarrow{FP}} \right|
\label{acc}
\end{equation}

\begin{equation}
	Precision = \left| \frac{\overrightarrow{T P}}{\overrightarrow{T P}+\overrightarrow{F P}} \right|
 \label{Prec}
\end{equation}

\begin{equation}
	F1 = \left(2 \times \frac{ { prec } \times r e c}{ { prec }+r e c}\right)
 \label{f1}
\end{equation}

\begin{equation}
	Recall = \left| \frac{\overrightarrow{T P}}{\overrightarrow{T P}+\overrightarrow{F N}} \right|
 \label{rec}
\end{equation}

\subsection{Case Western Reserve University Bearing Dataset}
\begin{figure}[htbp]
    \centering
    \subfigure[Original Signal]{\includegraphics[width=0.22\textwidth]{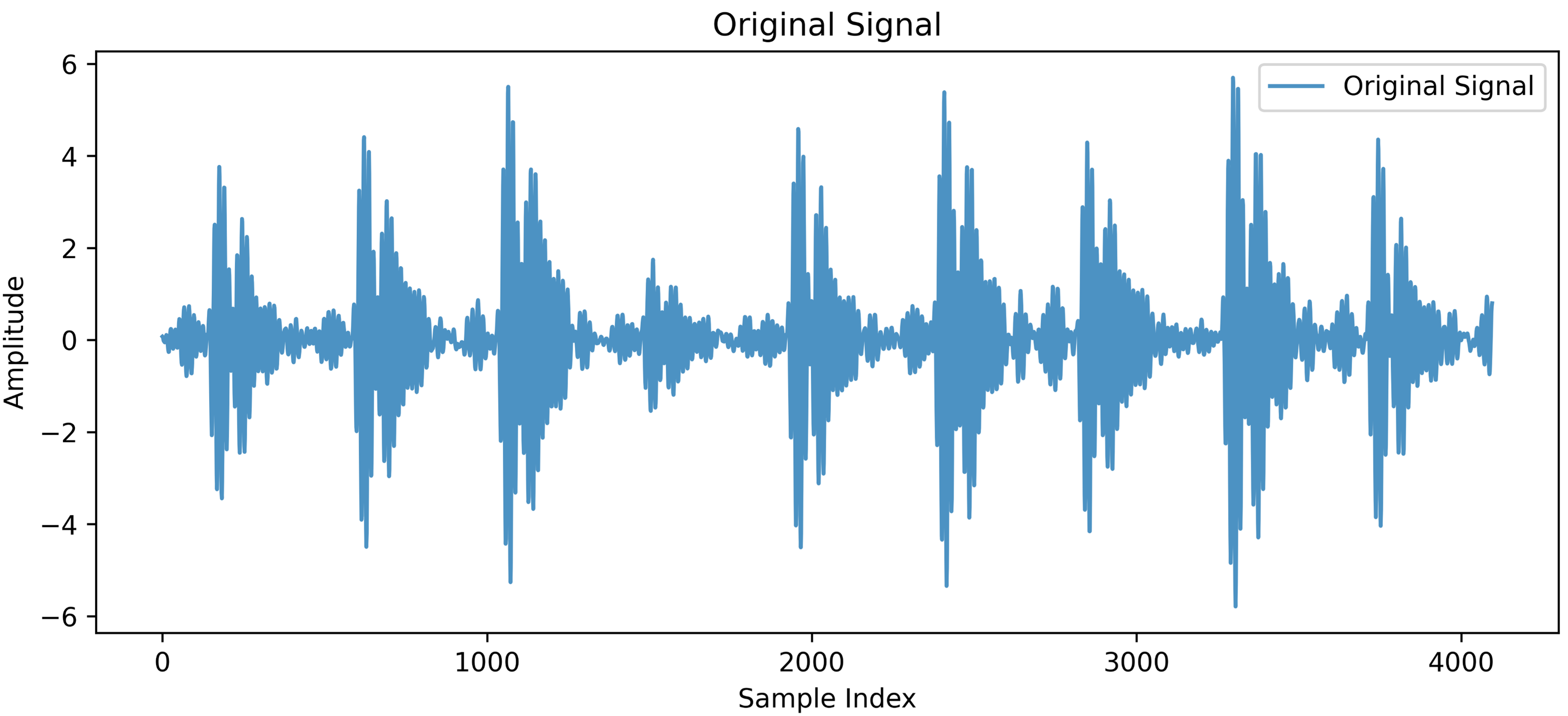}}
    \subfigure[Noisy Signal (SNR=-4dB)]{\includegraphics[width=0.22\textwidth]{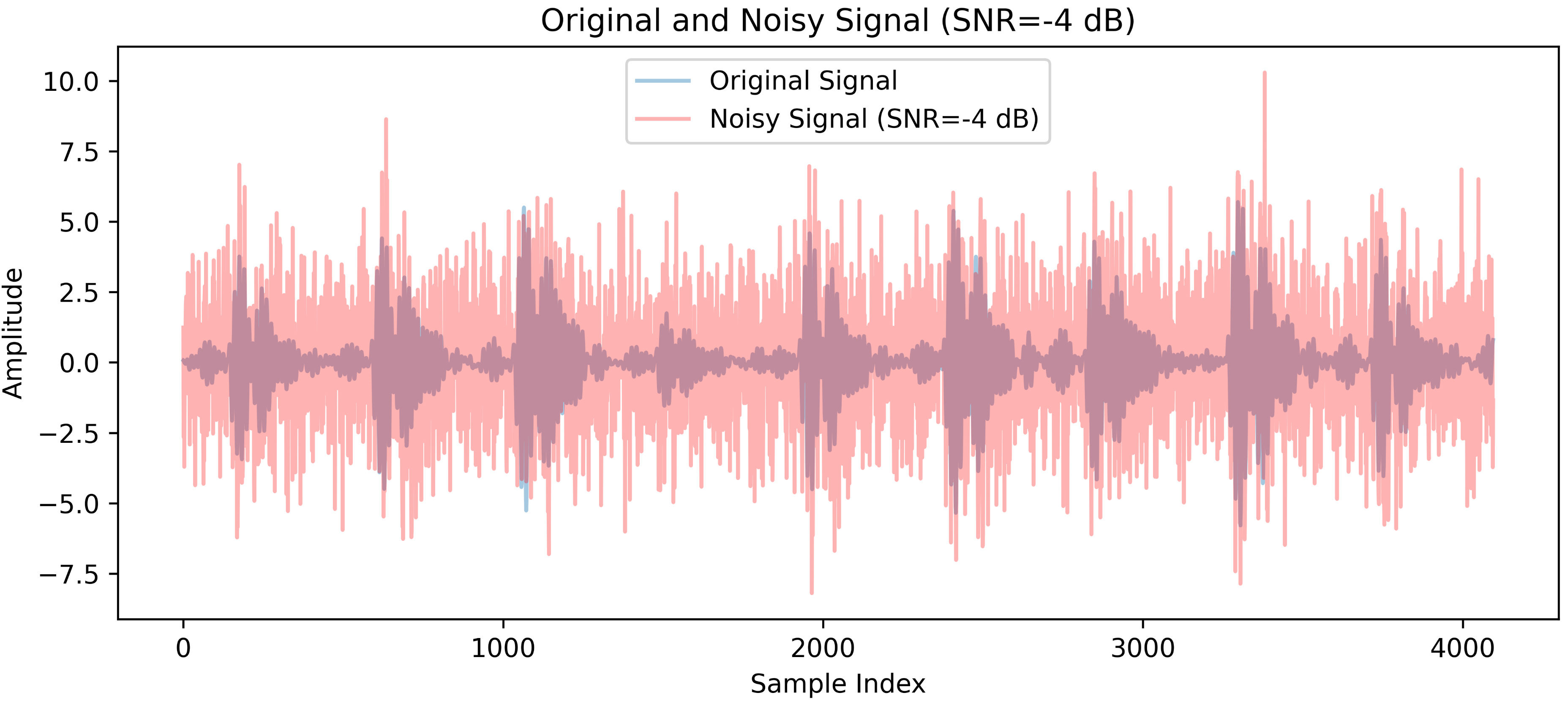}}
    \hspace{0.0\textwidth}
    \caption{The comparison of the original signal and the noise signal.}
    \label{fig_01}
\end{figure}

SKF rolling bearing of model 6205-2RS is employed in Case Western Reserve University (CWRU) bearing dataset. Vibration signals of the rolling bearing are collected at a sampling frequency of 48 kHz. Single-point faults are induced in the ball, inner raceway, and outer raceway, with fault diameters of 0.178 $mm$, 0.356 $mm$ and 0.532 $mm$, respectively. Therefore, the dataset contains 10 types of labels (9 for faults and 1 for normal). In order to facilitate feature extraction, the collected signals are subjected to min-max normalization to limit the preprocessed data within a certain range. The dataset is noise-free, and the current methods for solving fault diagnosis under noise environment are generally implemented by adding noise to the pure dataset. Therefore, the Gaussian white noise is used to simulate the interference in real scenes to verify the ability of the proposed method. The comparison of the original signal and the noise signal is shown in Fig. \ref{fig_01}. The equation for the SNR is shown in (\ref{SNR}):
\begin{equation}
S N R = 10 \log \left(\frac{P_{signal}}{P_{noise}}\right)
\label{SNR}
\end{equation}
where $P_{signal}$ and $P_{noise}$ represent the effective power of the signal and noise respectively.

\subsubsection{Comparison with traditional algorithms}
\begin{figure}[htbp]
    \centering
    \subfigure[Accuracy]{\includegraphics[width=0.23\textwidth]{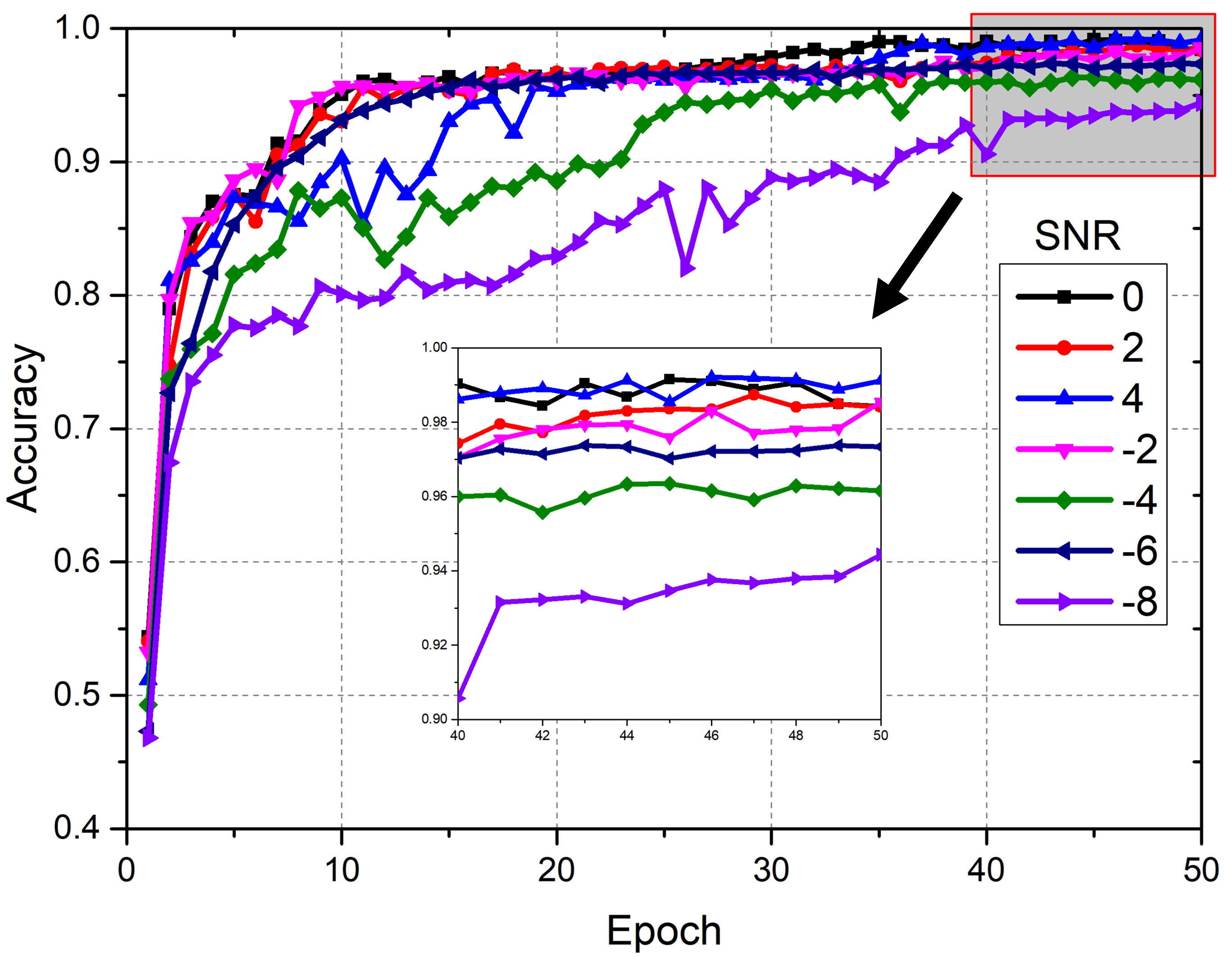}}
    \hspace{0.0\textwidth}
    \subfigure[Precision]{\includegraphics[width=0.23\textwidth]{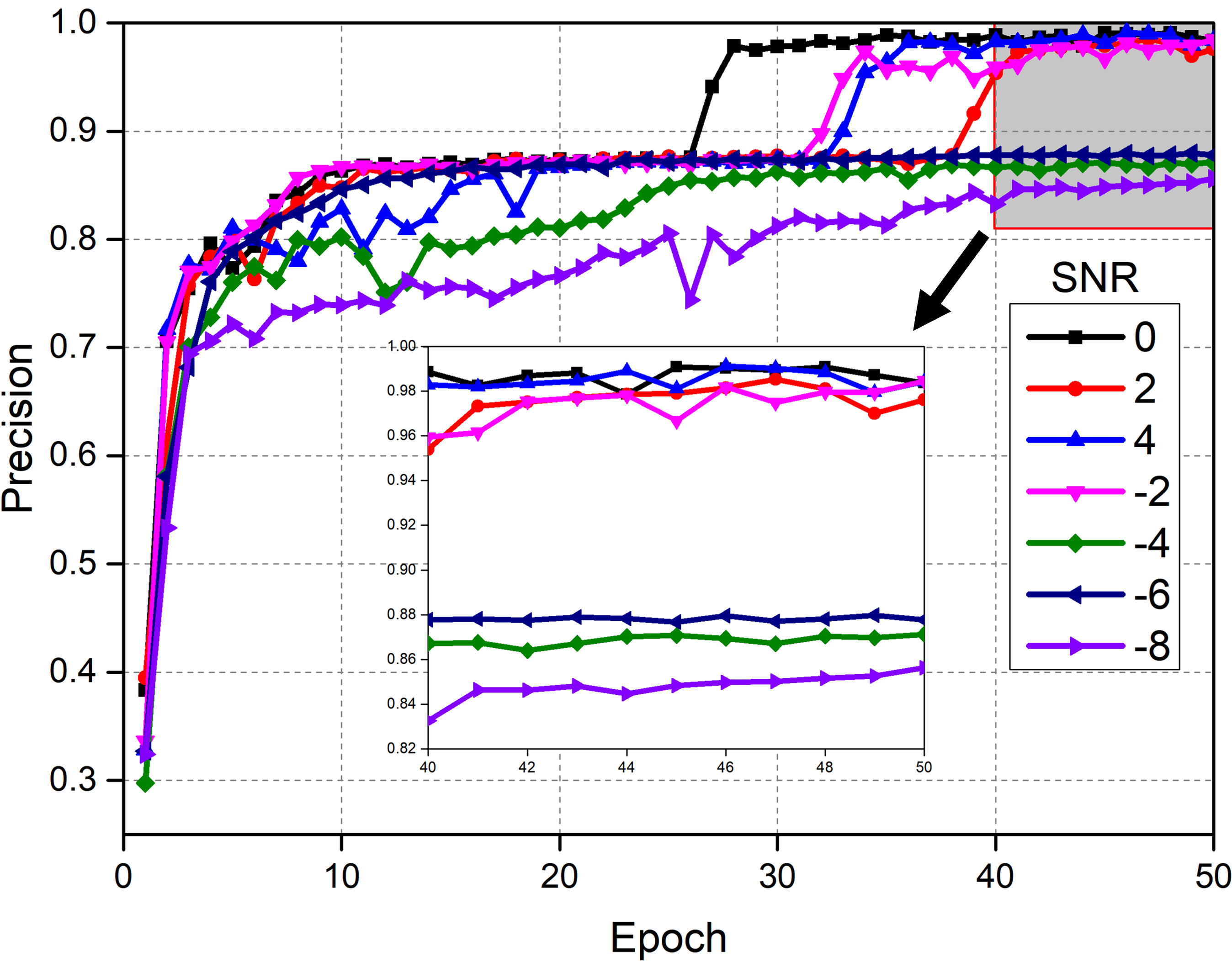}}\\
    \subfigure[F1]{\includegraphics[width=0.23\textwidth]{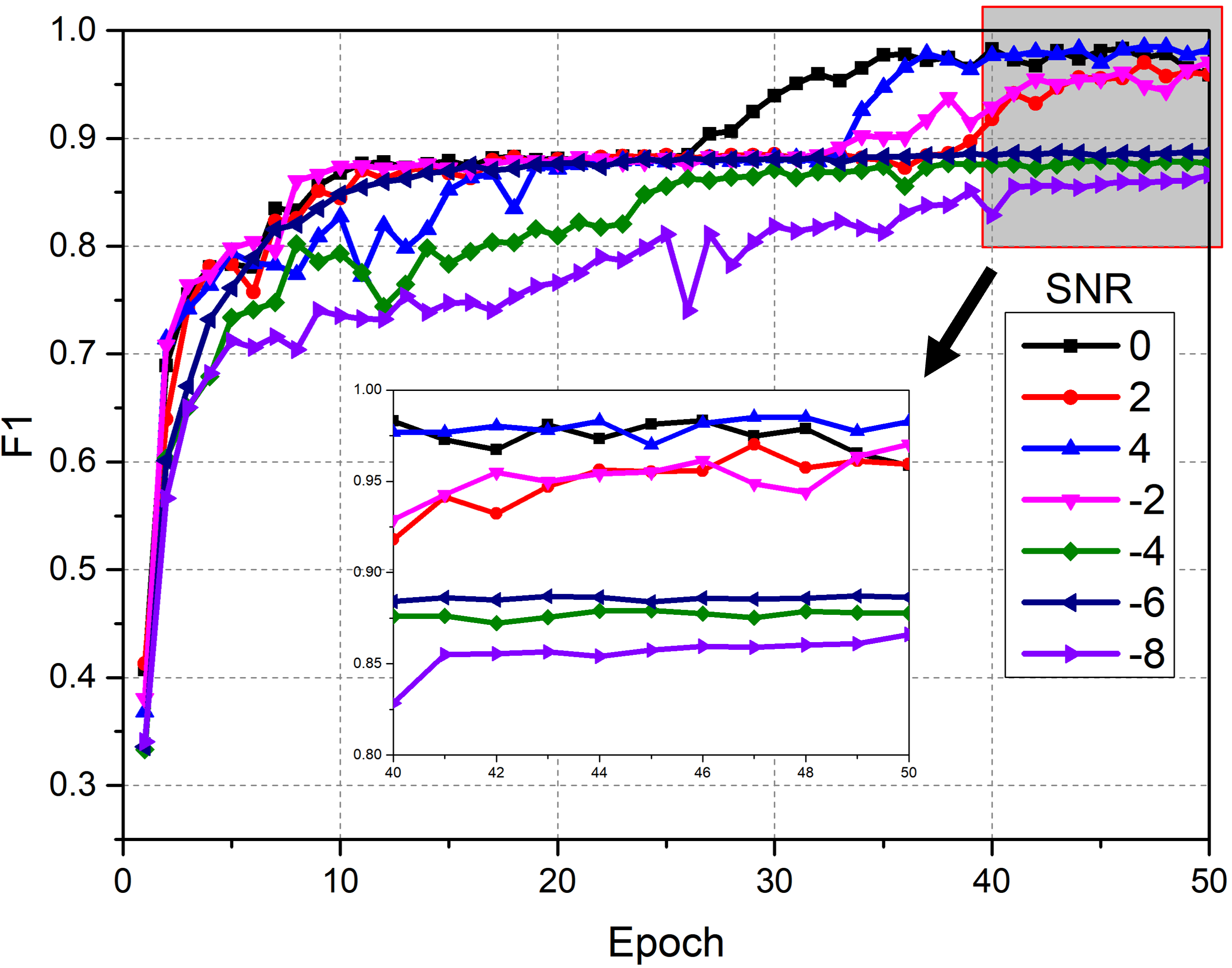}}
    \hspace{0.0\textwidth}
    \subfigure[Recall]{\includegraphics[width=0.23\textwidth]{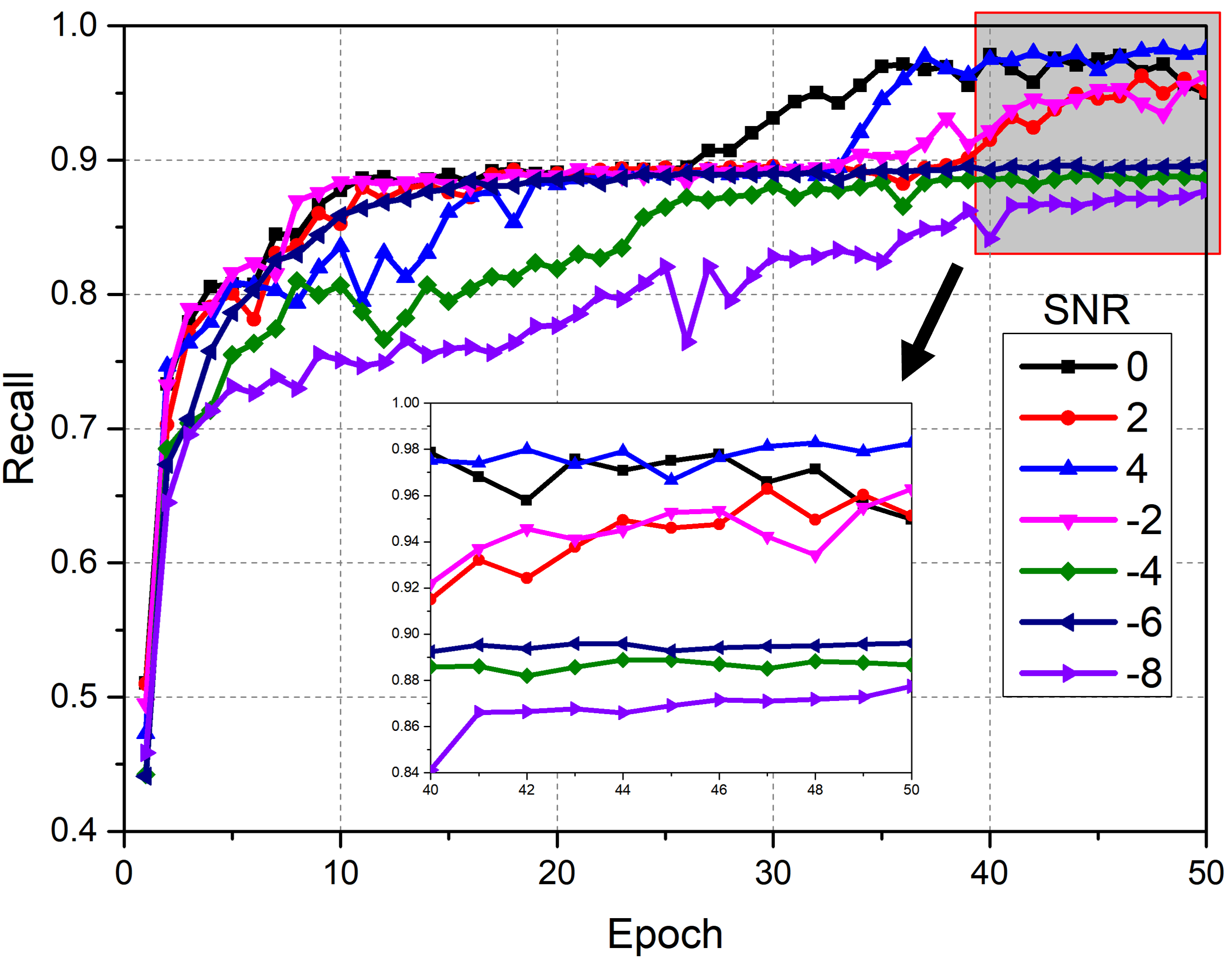}}
    \caption{Diagnostic results of the proposed TDANet under different SNRs.}
    \label{fig_02}
\end{figure}

In order to verify the effectiveness of the proposed TDANet, 5 algorithms were used for comparative experiments, including 1D-CNN, Pretrained AlexNet (2D-CNN), CNN-SVM (2D-CNN), CNN-ELM (2D-CNN) and CNN -LSTM (2D-CNN). The original vibration signal data was divided into 4096 segments and shuffled (a total of 9289), 80\% of the data was used as training data, and the remaining data was used as test data. The variation range of SNR in the experiment was set from -8 to 4 dB, with a step size of 2. The number of iterations of the experiment was set to 50. As the iteration proceeds, the changes in the 4 evaluation indicators for fault diagnosis using TDANet are shown in Figure \ref{fig_02}. The diagnostic accuracy of TDANet is obviously affected by the noise in the data, but it still has excellent diagnostic capabilities for signals under different SNRs and ensures convergence performance.

The comparison results with other algorithms are shown in Table \ref{exp:CC}. 1D CNN is limited by feature extraction capability, and its diagnostic performance is much lower than that of other algorithms. Algorithms based on 2D CNN have the ability to capture more complex features. However, Pretrained AlexNet, CNN-ELM and CNN-LSTM have significantly reduced diagnostic accuracy when the SNR is -8 dB. TDANet can maintain excellent diagnostic performance under low SNR. It has 94.56\% and 97.47\% classification accuracy at -8dB and -6dB respectively, which are 28.31\% and 21.72\% higher than the suboptimal algorithm CNN-SVM. In all SNR experiments, TDANet achieved the best diagnostic accuracy, which verifies the effectiveness of the method proposed in this paper.
\begin{table}[]
\centering
\caption{The comparison experiments on CWRU.}
\resizebox{0.50\textwidth}{!}{
\begin{tabular}{cc|ccccccc|c}
\hline
\hline
\multicolumn{1}{c}{\multirow{2}{*}{Algorithm}}       & & \multicolumn{7}{c|}{SNR}                              & \multirow{2}{*}{Average} \\ \cline{3-9}
\multicolumn{2}{c|}{}                                              & -8    & -6    & -4    & -2    & 0     & 2     & 4     &                      \\ \hline
\multicolumn{1}{c|}{\multirow{4}{*}{1D CNN}}              & Acc    & 14.72 & 18.45 & 26.31 & 33.97 & 39.11 & 43.25 & 48.49 & 32.04                \\
\multicolumn{1}{c|}{}                                     & Pre    & 14.24 & 15.37 & 23.83 & 30.86 & 41.58 & 47.18 & 55.96 & 32.72                \\
\multicolumn{1}{c|}{}                                     & Recall & 14.45 & 18.25 & 26.47 & 34.49 & 39.46 & 43.93 & 49.24 & 32.33                \\
\multicolumn{1}{c|}{}                                     & F1     & 14.25 & 15.95 & 24.59 & 31.67 & 38.39 & 42.05 & 48.24 & 30.74                \\ \hline
\multicolumn{1}{c|}{\multirow{4}{*}{Pretrained AlexNet}}      & Acc    & 31.25 & 48.29 & 55.95 & 69.15 & 91.03 & 72.28 & 95.97 & 66.27                \\
\multicolumn{1}{c|}{}                                     & Pre    & 38.43 & 54.57 & 64.84 & 72.61 & 92.73 & 84.59 & 96.31 & 72.01                \\
\multicolumn{1}{c|}{}                                     & Recall & 32.17 & 48.88 & 56.68 & 69.84 & 91.51 & 72.57 & 96.19 & 66.83                \\
\multicolumn{1}{c|}{}                                     & F1     & 26.86 & 47.58 & 54.45 & 67.71 & 91.01 & 70.86 & 96.07 & 64.94                \\ \hline
\multicolumn{1}{c|}{\multirow{4}{*}{CNN-SVM}}     & Acc    & 66.25 & 75.75 & 78.00 & 88.13 & 93.13 & 94.88 & 95.88 & 84.57                \\
\multicolumn{1}{c|}{}                                     & Pre    & 67.23 & 75.64 & 77.70 & 87.91 & 92.90 & 94.70 & 95.58 & 84.52                \\
\multicolumn{1}{c|}{}                                     & Recall & 66.57 & 75.44 & 77.52 & 87.76 & 93.20 & 94.71 & 95.60 & 84.4                 \\
\multicolumn{1}{c|}{}                                     & F1     & 66.64 & 75.42 & 77.46 & 87.78 & 93.00 & 94.66 & 95.58 & 84.36                \\ \hline
\multicolumn{1}{c|}{\multirow{4}{*}{CNN-ELM}}     & Acc    & 18.04 & 29.64 & 42.24 & 61.90 & 79.54 & 88.00 & 93.95 & 59.04                \\
\multicolumn{1}{c|}{}                                     & Pre    & 19.29 & 30.29 & 43.01 & 62.89 & 80.63 & 88.62 & 94.42 & 59.88                \\
\multicolumn{1}{c|}{}                                     & Recall & 18.34 & 30.40 & 42.88 & 63.18 & 80.80 & 88.56 & 94.42 & 59.8                 \\
\multicolumn{1}{c|}{}                                     & F1     & 18.66 & 30.17 & 42.58 & 62.49 & 80.12 & 88.52 & 94.36 & 59.56                \\ \hline
\multicolumn{1}{c|}{\multirow{4}{*}{CNN-LSTM}}            & Acc    & 17.54 & 30.85 & 40.63 & 49.09 & 56.45 & 70.16 & 81.25 & 49.42                \\
\multicolumn{1}{c|}{}                                     & Pre    & 18.27 & 31.05 & 41.37 & 49.83 & 57.26 & 70.27 & 81.26 & 49.9                 \\
\multicolumn{1}{c|}{}                                     & Recall & 17.80 & 31.70 & 41.41 & 50.46 & 57.92 & 71.55 & 82.25 & 50.44                \\
\multicolumn{1}{c|}{}                                     & F1     & 17.93 & 31.20 & 41.13 & 50.05 & 57.49 & 70.68 & 81.50 & 50                   \\ \hline
\multicolumn{1}{c|}{\multirow{4}{*}{Proposed TDANet}}     & Acc    & 94.56 & 97.47 & 96.66 & 97.79 & 99.25 & 98.76 & 99.35 & 97.69                \\
\multicolumn{1}{c|}{}                                     & Pre    & 85.75 & 87.88 & 87.44 & 98.31 & 99.13 & 97.3  & 98.62 & 93.49                \\
\multicolumn{1}{c|}{}                                     & Recall & 87.86 & 89.71 & 89.14 & 94.37 & 97.76 & 96.27 & 98.63 & 93.39                \\
\multicolumn{1}{c|}{}                                     & F1     & 86.71 & 88.75 & 88.2  & 95.55 & 98.36 & 96.71 & 98.63 & 93.27                \\ \hline \hline
\end{tabular}
\label{exp:CC}
}
\end{table}

\subsubsection{Ablation analysis}
To demonstrate the effectiveness of STFT, we set up 7 sets of experiments with SNR ranging from -8 to 4 dB and a step size of 2. The hyperparameter $k$ (number of stacked images) ranges from 2 to 8 with a step size of 2. The experimental results are shown in Table \ref{table:expofFC}. Applying STFT decomposition can effectively improve the diagnostic accuracy, especially when the $k$ is large, which reveals the powerful ability of STFT to handle non-stationary signals. As the $k$ increases, the model can capture more detailed information, which is beneficial to providing more accurate diagnostic results. When the $k$ reaches 8, the average accuracy of the model in 7 sets of experiments is 97.69\%, which is an improvement of 5.19\% compared to applying FFT.

\begin{table}[htbp]
\caption{The diagnostic accuracy (\%) of different decomposition methods on CWRU.}
\resizebox{0.50\textwidth}{!}{
\begin{tabular}{c|c|ccccccc|c}
\hline
\hline
\multirow{2}{*}{Algorithm} & \multirow{2}{*}{$k$} & \multicolumn{7}{c|}{SNR}                              & \multirow{2}{*}{Average} \\ \cline{3-9}
                                      &                    & -8    & -6    & -4    & -2    & 0     & 2     & 4     &                      \\ \hline
\multirow{4}{*}{FFT}                  & 2                  & 89.71 & 90.19 & 91.11 & 90.41 & 93.05 & 94.34 & 91.81 & 91.52                \\
                                      & 4                  & 92.81 & 89.33 & 88.2  & 93.21 & 93.16 & 92.4  & 96.28 & 92.2                 \\
                                      & 6                  & 90.52 & 91.65 & 92.35 & 94.23 & 96.23 & 94.72 & 92.73 & 93.2                 \\
                                      & 8                  & 80.87 & 93.64 & 93.86 & 94.07 & 95.26 & 95.1  & 94.78 & 92.51                \\ \hline
\multirow{4}{*}{STFT}                 & 2                  & 88.42 & 89.44 & 88.58 & 91.49 & 92.13 & 93.97 & 91.06 & 90.73 ($\downarrow$ 0.52)                \\
                                      & 4                  & 94.45 & 94.07 & 95.37 & 94.83 & 94.23 & 96.55 & 94.94 & 94.92 ($\uparrow$ 2.72)                \\
                                      & 6                  & 94.45 & 96.12 & 96.28 & 97.14 & 96.5  & 97.74 & 99.03 & 96.75 ($\uparrow$ 3.55)                \\
                                      & \textbf{8}      & \textbf{94.56} & \textbf{97.47} & \textbf{96.66} & \textbf{97.79} & \textbf{99.25} & \textbf{98.76} & \textbf{99.35} & \textbf{97.69 ($\uparrow$ 5.19)}                \\ \hline \hline
\end{tabular}
}
\label{table:expofFC}
\end{table}

\begin{figure}[htbp]
    \centering
    \subfigure[Accuracy]{\includegraphics[width=0.23\textwidth]{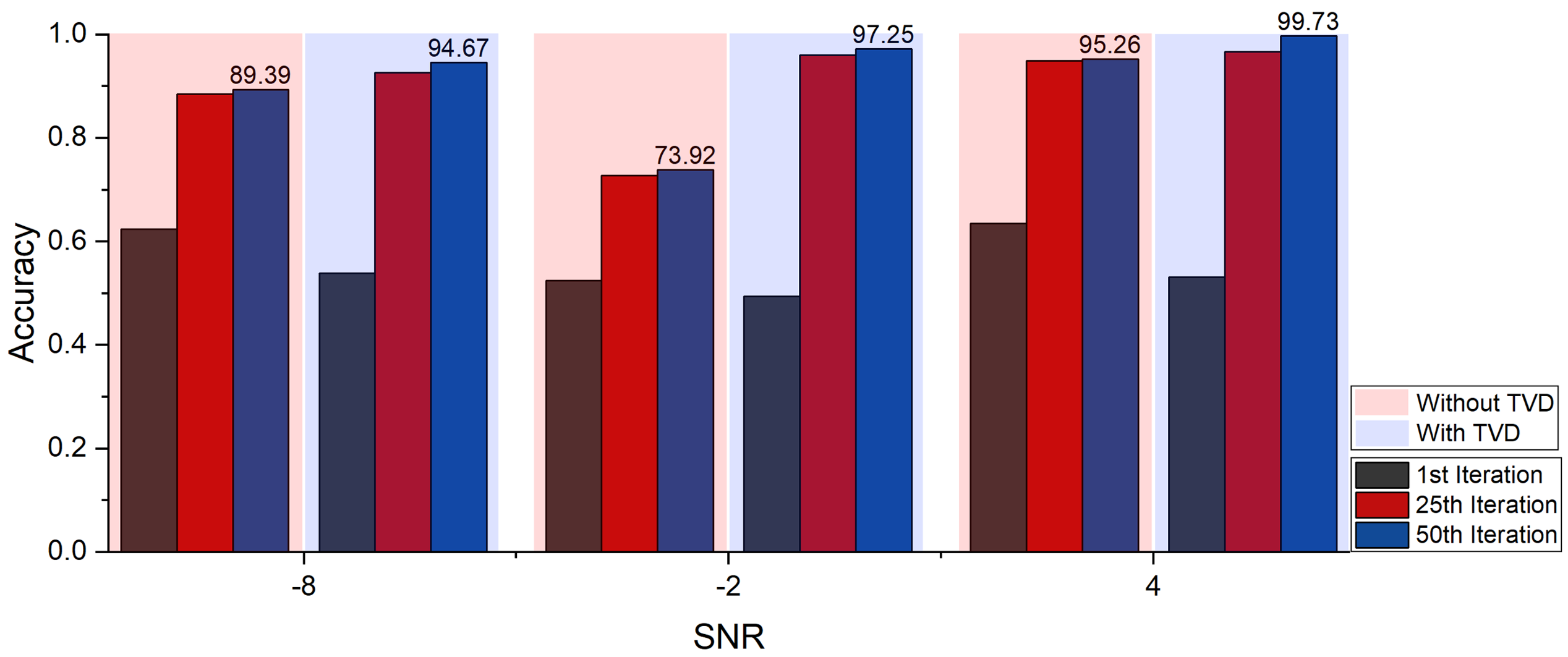}}
    \hspace{0.0\textwidth}
    \subfigure[Precision]{\includegraphics[width=0.23\textwidth]{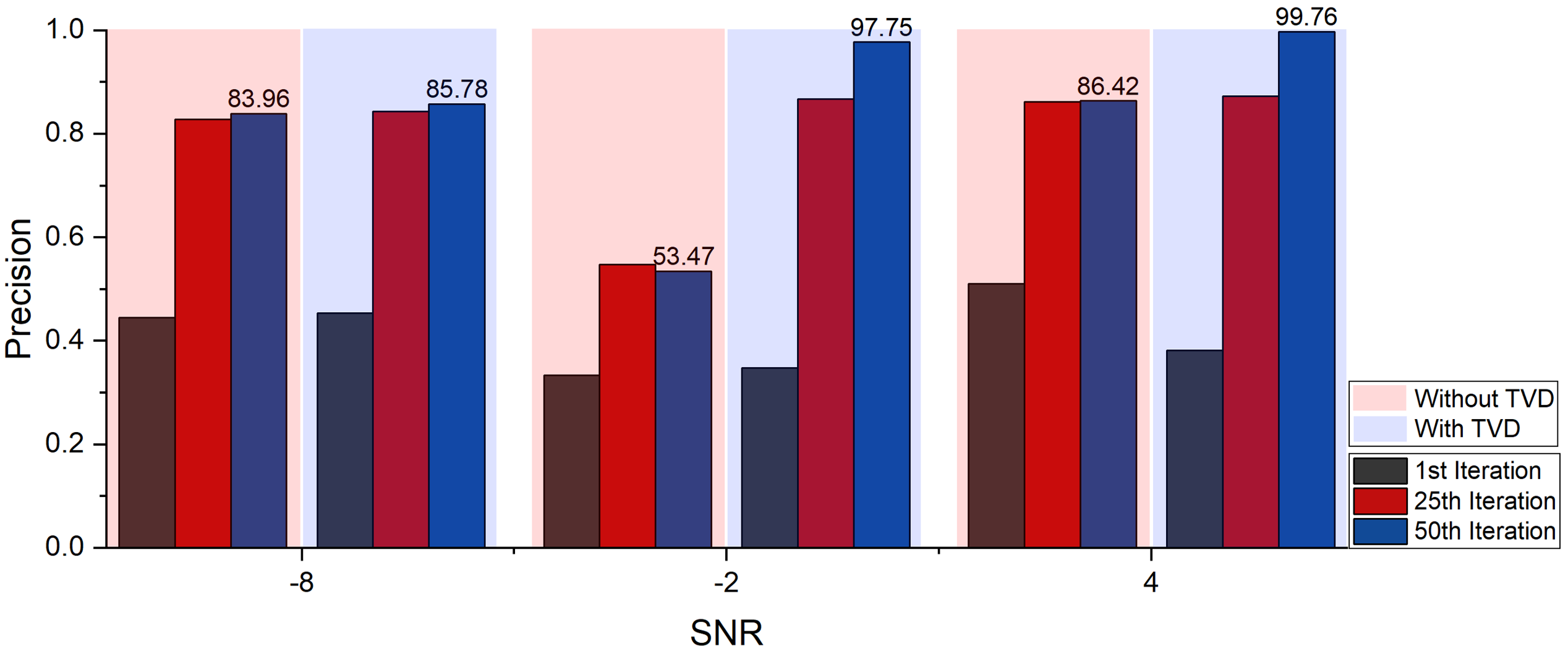}}\\
    \subfigure[F1]{\includegraphics[width=0.23\textwidth]{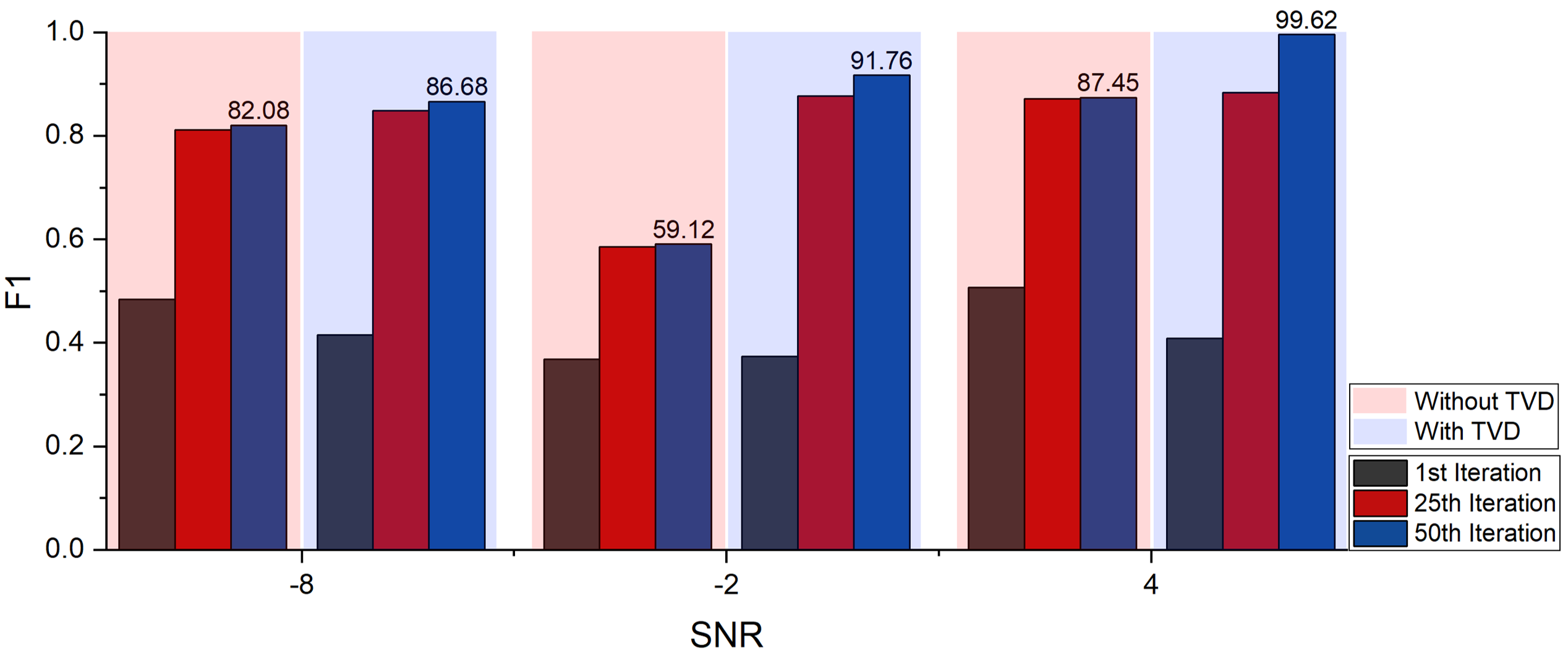}}
    \hspace{0.0\textwidth}
    \subfigure[Recall]{\includegraphics[width=0.23\textwidth]{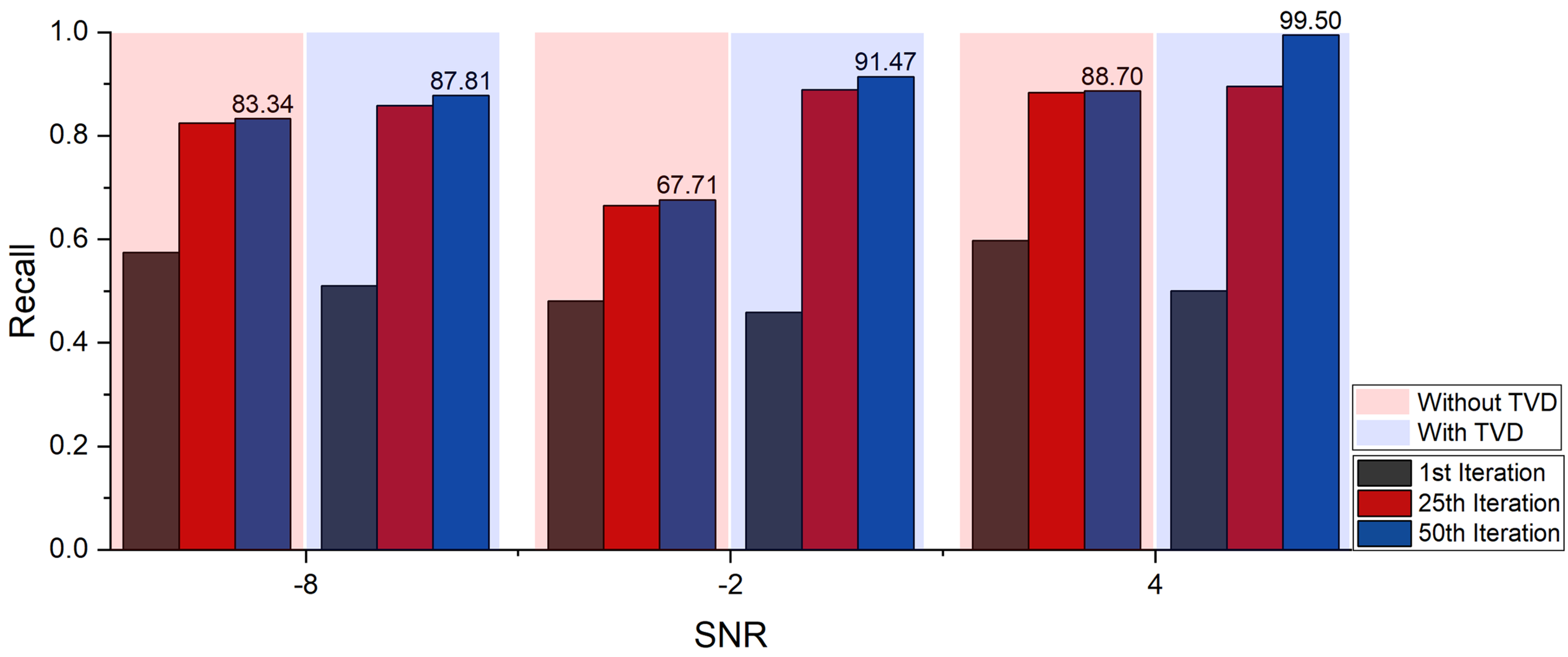}}
    \caption{The ablation analysis results of TVD module.}
    \label{fig_03}
\end{figure}

\begin{figure}[htbp]
    \centering
    \subfigure[Accuracy]{\includegraphics[width=0.23\textwidth]{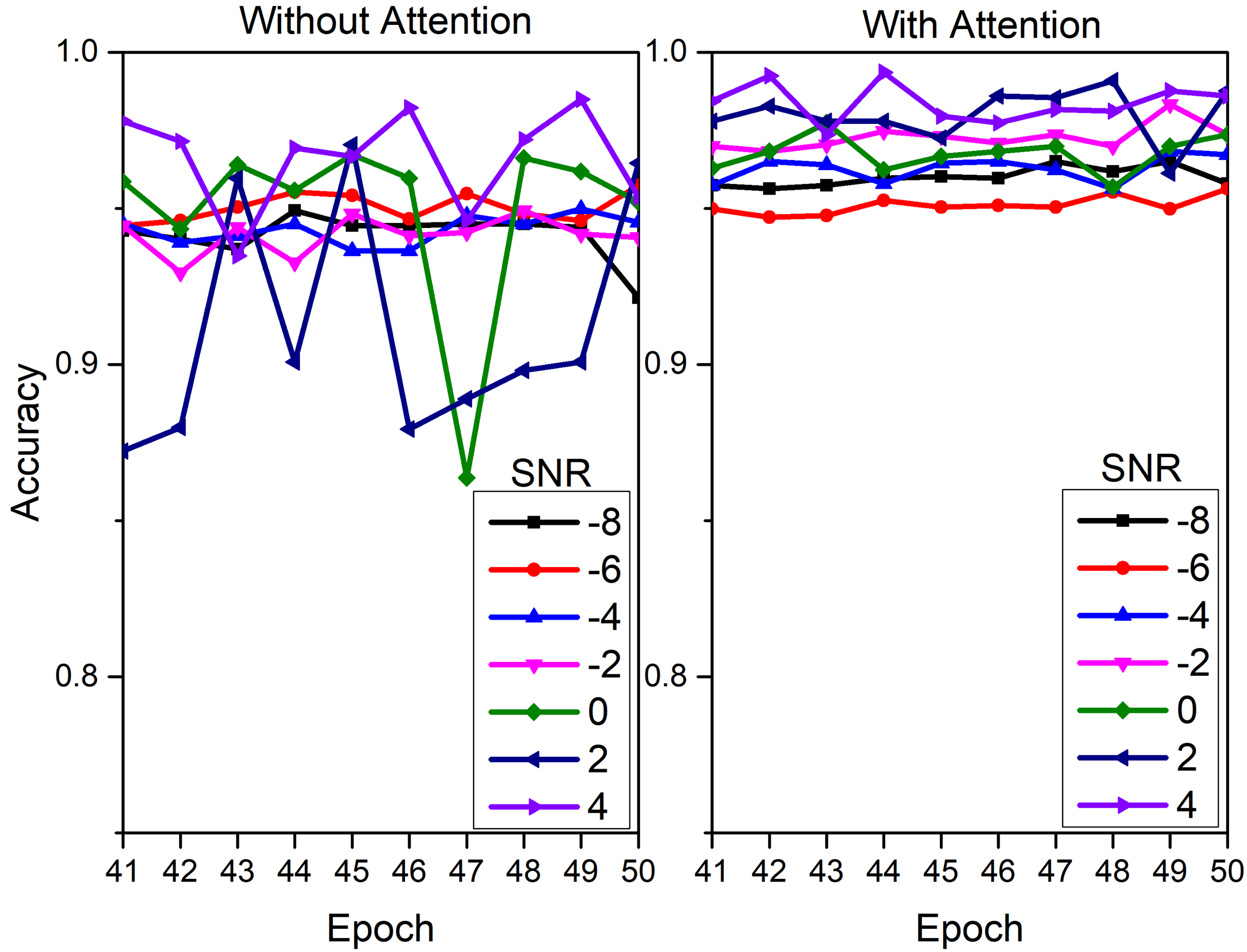}}
    \hspace{0.0\textwidth}
    \subfigure[Precision]{\includegraphics[width=0.23\textwidth]{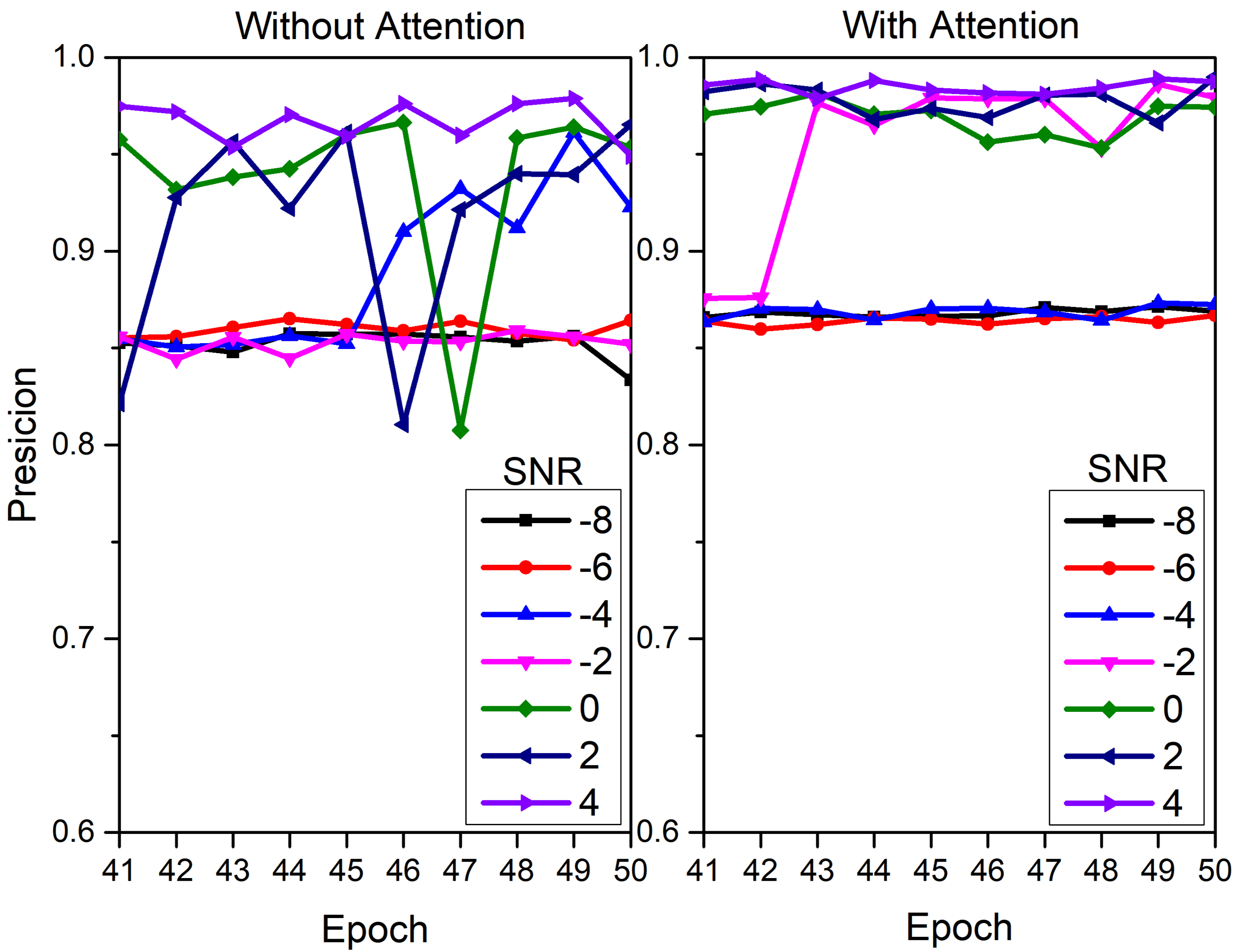}}\\
    \subfigure[F1]{\includegraphics[width=0.23\textwidth]{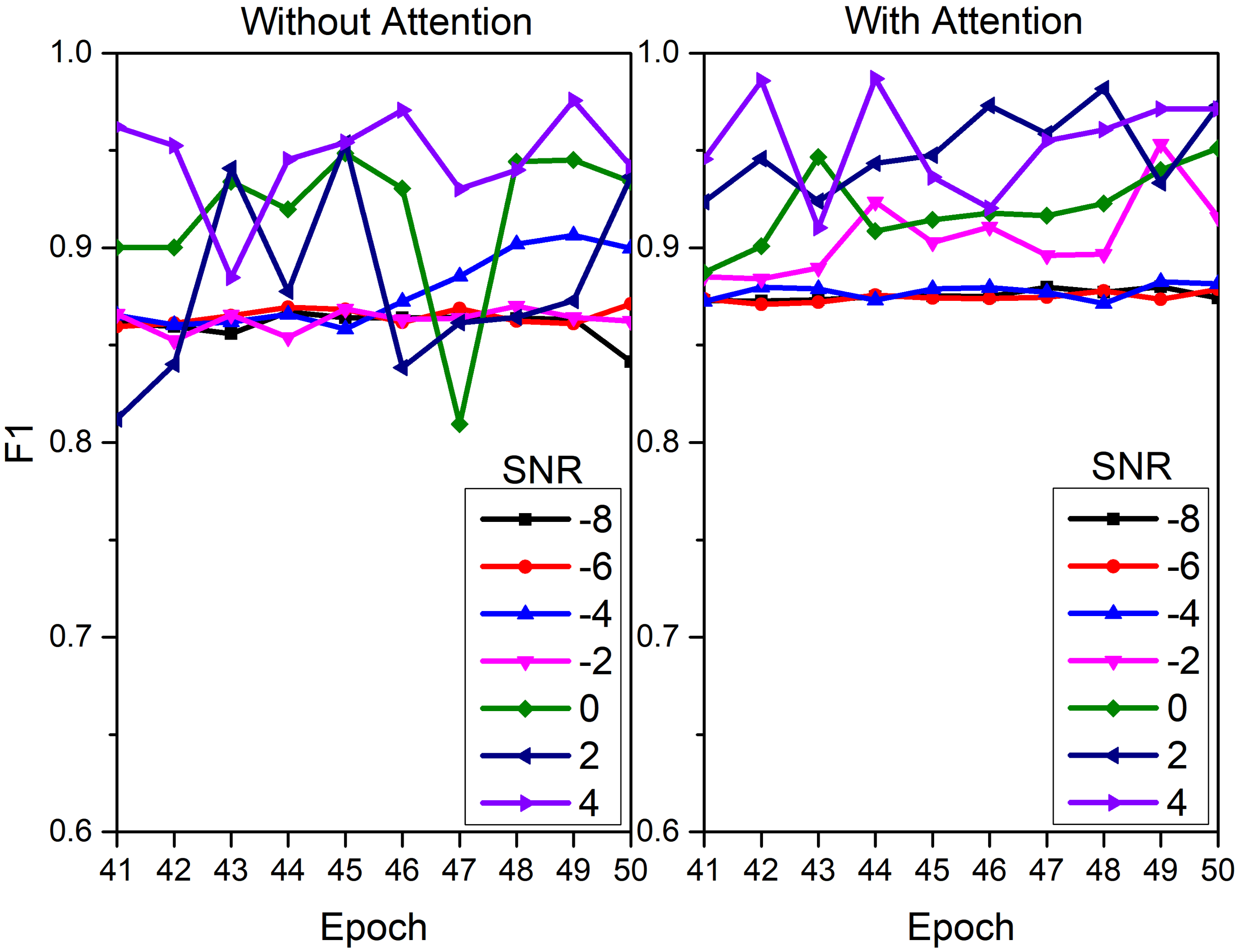}}
    \hspace{0.0\textwidth}
    \subfigure[Recall]{\includegraphics[width=0.23\textwidth]{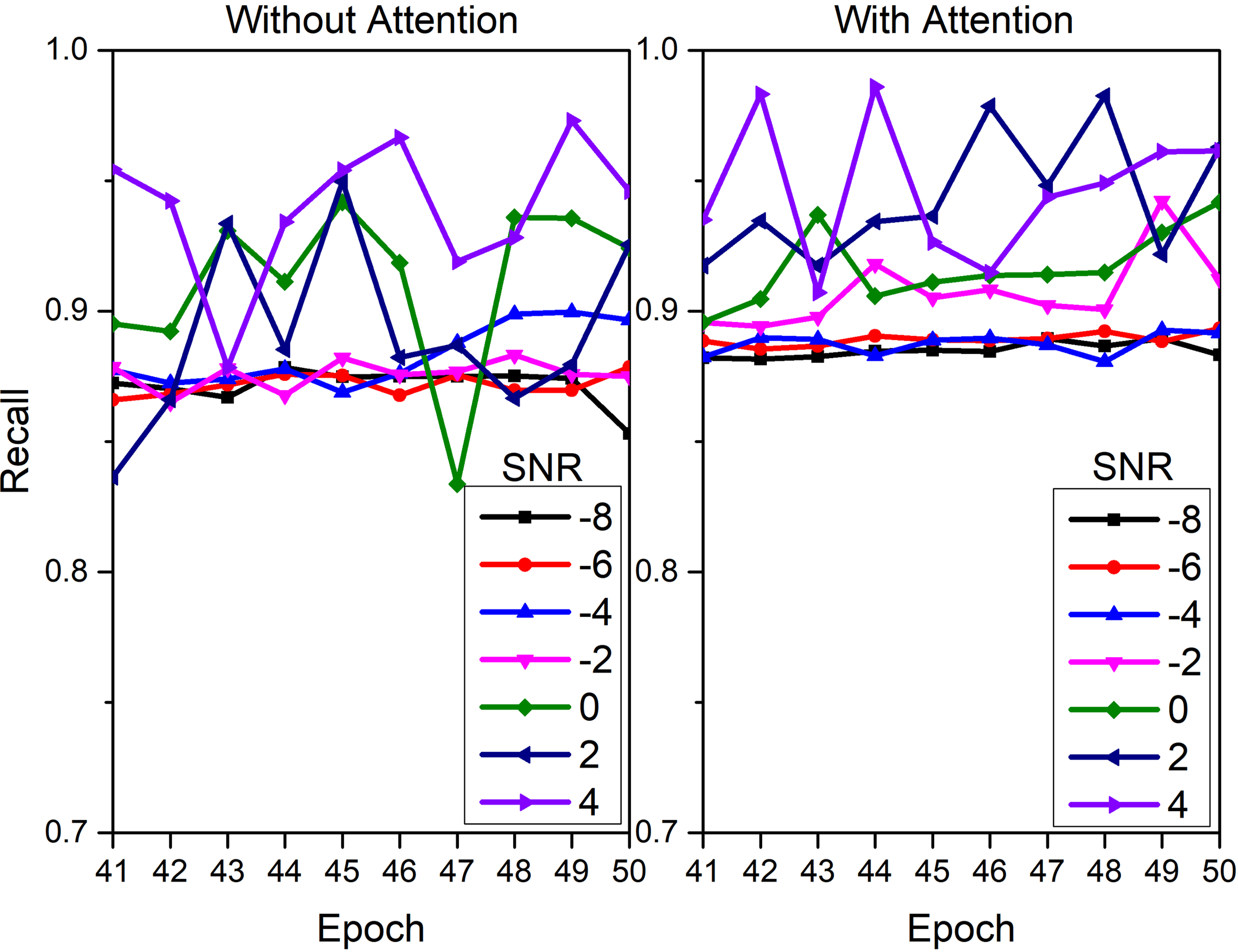}}
    \caption{The ablation analysis results of MAF module.}
    \label{fig_04}
\end{figure}

The ablation analysis results of TVD and MAF modules are shown in Figures \ref{fig_03} and \ref{fig_04} respectively. For the TVD module, we set up 3 sets of experiments at SNRs of -8, -2 and 4 dB. It can be clearly observed that adding the TVD module can effectively improve the diagnostic accuracy of the model, especially when the SNR is -2 dB (23.33\% improvement in accuracy). For the MAF module, we set up 7 sets of experiments with SNR ranging from -8 to 4 dB and a step size of 2. Figure \ref{fig_04} shows the results from the 41st iteration to the 50th iteration. It can be found that by adding the MAF module, the model can achieve higher diagnostic accuracy and make the final accuracy change curve smoother.

\subsubsection{Comparison of training efficiency}
\begin{figure}[htbp]
\centering
\includegraphics[clip,width = 0.9
\linewidth]{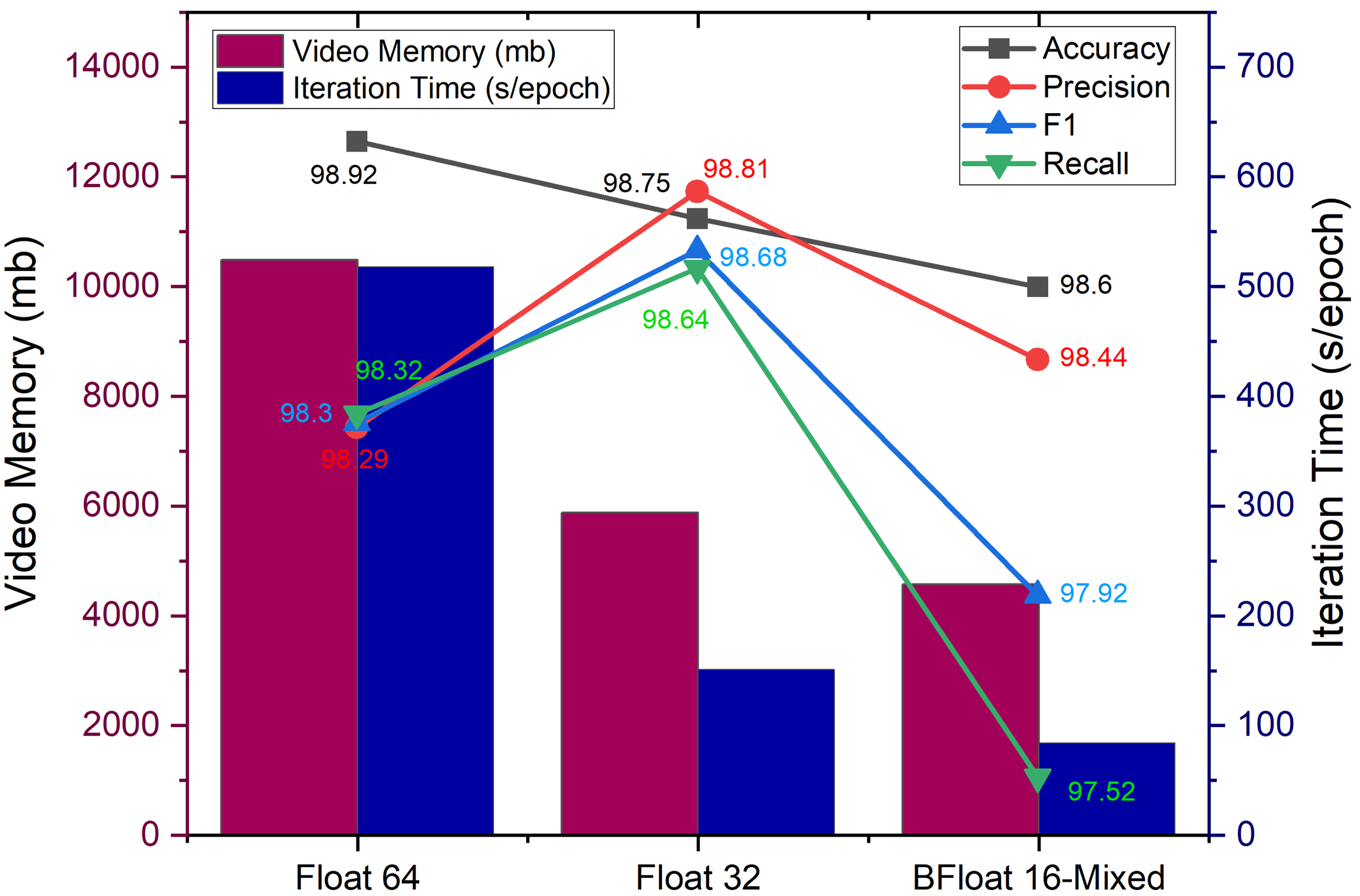}
\caption{Results of different calculation strategies in terms of diagnostic accuracy, video memory consumption and iteration time.}
\label{Fig:FloatC}
\end{figure}

The optimization strategy of mixed precision calculation is applied to the training of TDANet. The results of the model trained with two precisions of Float 32 and Float 64 in terms of diagnostic accuracy, video memory and iteration time are shown in Figure \ref{Fig:FloatC}. Applying mixed-precision calculations will slightly reduce the diagnostic accuracy of the model (0.12\% of the Float 64), greatly reduce memory consumption (43.65\% of Float 64) and iteration time (16.22\% of Float 64), and improve the efficiency of training models. This is not a task-dependent strategy and can be embedded in the training of other deep learning models.

\subsection{Real Aircraft Sensor Fault Dataset}
\subsubsection{Dataset Description} 
\begin{figure}[htbp]
	\centering
	\includegraphics[width =0.8\linewidth,trim=100 0 100 0]{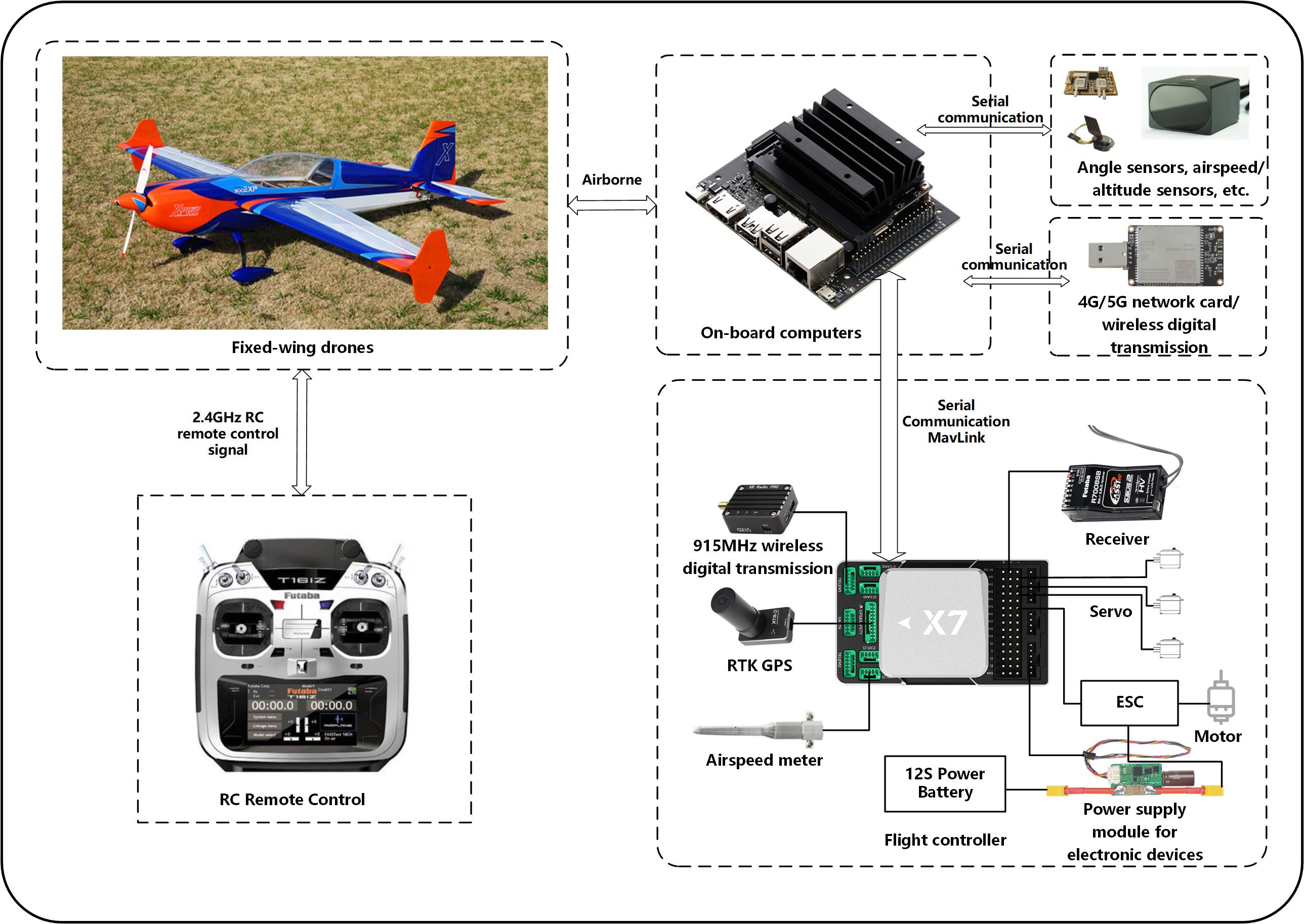}
	\caption{Configurations of the UAV adopted in this paper.}
	\label{fig:Overall_Design}
\end{figure}

\begin{figure}[htbp]
    \centering
    \subfigure[Original Signal]{\includegraphics[width=0.22\textwidth]{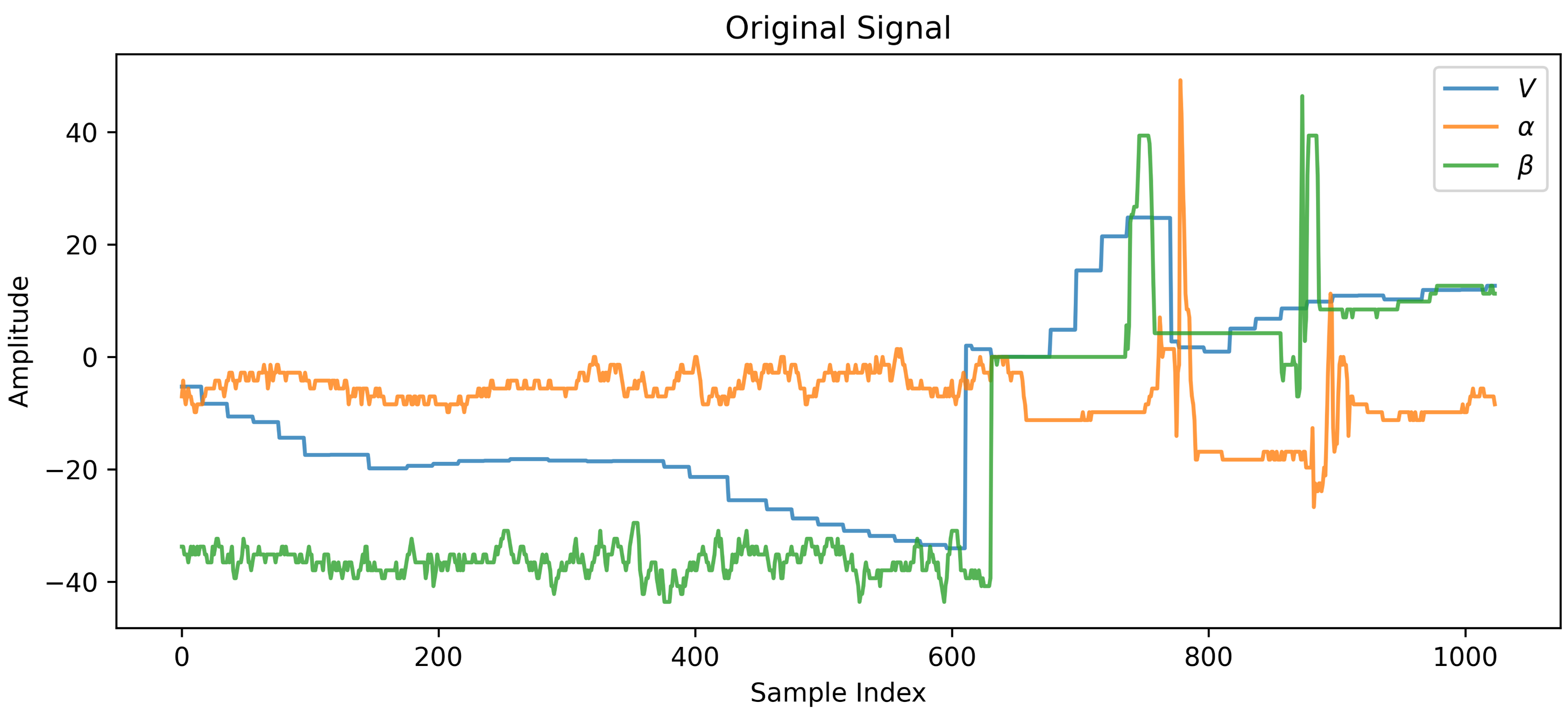}}
    \hspace{0.0\textwidth}
    \subfigure[Noisy Signal (SNR=6dB)]{\includegraphics[width=0.22\textwidth]{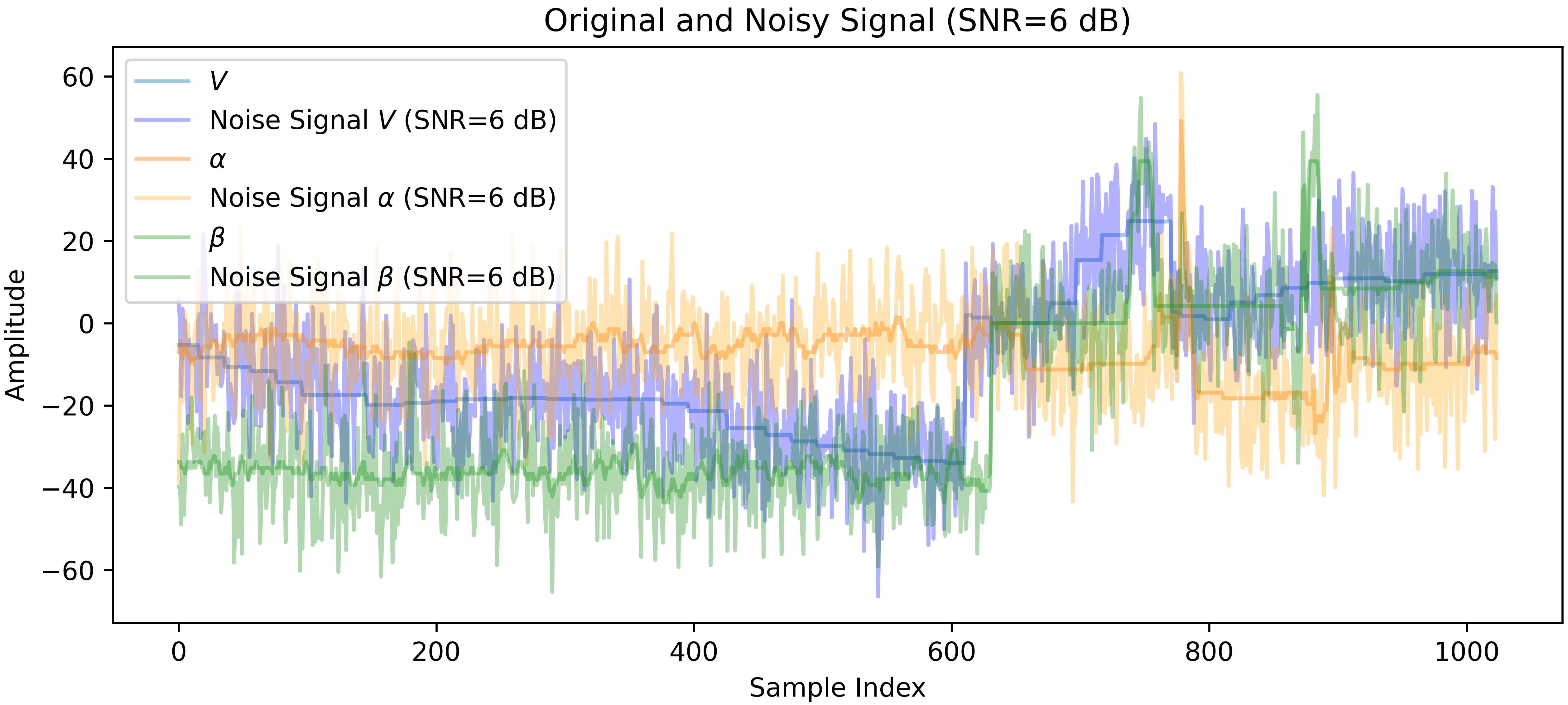}}
    \caption{The comparison of the original signal and the noise signal.}
    \label{fig_05}
\end{figure}

In this study, we conducted real aircraft data collection using a 78-inch EXTRA 300 NG fixed-wing Unmanned Aerial Vehicle (UAV) equipped with the CUAV X7+ PRO flight control system, as depicted in Figure \ref{fig:Overall_Design}. This UAV is equipped with a range of sensors, including the CUAV C-RTK 9P GPS, SMV-1 non-contact Hall principle angle measuring vane, and ADM800 altitude/airspeed meter. Notably, many aviation accidents have been attributed to the Pitot tube becoming obstructed, leading to airspeed-related issues. Consequently, drift faults (manifesting as measurement loss) are taken into consideration for airspeed sensors ($V_m$). As for Angle of Attack (AOA, $\alpha_m$) and sideslip angle sensors ($\beta_m$), potential issues may arise from deflection vanes getting stuck or perturbed by external atmospheric conditions, giving rise to drift (constant bias) and additional noise faults. As detailed in Table \ref{fault_cases}, a total of 5 distinct fault cases are explored. By combining the normal data, we have established a real aircraft sensor fault dataset that encompasses 6 categories. Gaussian white noise with different SNRs (-4, 0, 4, 6, 8, 10, and 20 dB) was added to the signals, and some results are shown in Figure \ref{fig_05}.

\begin{table}[htbp]
    \centering
    \caption{Aircraft sensors fault cases adopted in this paper.}
    \resizebox{0.5\textwidth}{!}{
    \label{fault_cases}
        \scalebox{0.9}{
    \begin{tabular}{c|ccc}
    \hline
    \hline
    \textbf{Case}   &\textbf{Sensor}   &\textbf{Fault type}  &\textbf{Magnitude\textbf{*}}          \\
    \hline
    \textbf{5}   &$\beta_m$     &extra noise&$5^o\sim10^o$              \\
    \textbf{4}   &$\beta_m$     &drift&$\pm(5^o\sim10^o)$         \\
    \textbf{3}   &$\alpha_m$                &extra noise&$5^o\sim10^o$              \\
    \textbf{2}   &$\alpha_m$                &drift&$\pm(5^o\sim10^o)$         \\
    \textbf{1}   &$V_m$           &drift&$-(50\%\sim100\%)$         \\   
    \hline
    \textbf{0}   &\multicolumn{3}{c}{clean measurement with noises and disturbances, no fault}  \\
    \hline
    \hline
    \multicolumn{4}{l}{\textbf{*} Noise standard deviation and drift values defined in this column.}
    \end{tabular}
    }
    }
\end{table}

\subsubsection{Ablation analysis}
On the Real aircraft sensor fault dataset, we set up experiments with 7 SNR conditions, and the results are shown in Table \ref{tab:expofCF}. Applying STFT can effectively improve the diagnostic accuracy of the model, especially when the $k$ is 2 (a 3.52\% improvement compared to FFT). Different from the results of the CWRU dataset, when the $k$ value is too large (8), more interference information will be introduced, resulting in a decrease in the performance of the model. The ablation analysis results for TVD and MAF modules are shown in Table \ref{tab:expofDF}. It can be found that only adding the MAF module can bring a slight improvement in diagnostic accuracy to the model, especially at low SNR (-4). The role of the TVD module is even more significant, which verifies its effectiveness in performing signal processing in noisy environments.

\begin{table}[]
\caption{The diagnostic accuracy (\%) of different decomposition methods on Flight.}
\resizebox{0.50\textwidth}{!}{
\begin{tabular}{c|c|ccccccc|c}
\hline
\hline
\multirow{2}{*}{Algorithm} & \multirow{2}{*}{$k$} & \multicolumn{7}{c|}{SNR}                              & \multirow{2}{*}{Average} \\ \cline{3-9}
                                      &                    & -4    & 0    & 4    & 6    & 8     & 10     & 20     &                      \\ \hline
\multirow{4}{*}{FFT}                  & 2                  & 62.86 & 77.14 & 83.57 & 77.86 & 94.11 & 95.71 & 98.93 & 84.31                \\
                                      & 4                  & 67.32 & 59.11 & 78.93 & 81.43 & 94.11 & 88.57 & 99.82 & 81.33                \\
                                      & 6                  & 52.5  & 66.07 & 66.96 & 84.64 & 90.54 & 95.71 & 96.07 & 78.93                \\
                                      & 8                  & 54.11 & 68.04 & 75.54 & 88.75 & 90    & 93.57 & 94.11 & 80.59                \\ \hline
\multirow{4}{*}{STFT}                 & \textbf{2}      & \textbf{71.79} & \textbf{78.75} & \textbf{81.07} & \textbf{90.89} & \textbf{96.43} & \textbf{96.07} & \textbf{99.82} & \textbf{87.83 ($\uparrow$ 3.52)}   \\
                                      & 4                  & 57.68 & 63.57 & 83.39 & 95.54 & 91.07 & 91.61 & 99.82 & 83.24 ($\uparrow$ 1.91)                \\
                                      & 6                  & 55.54 & 63.75 & 80.89 & 87.86 & 88.57 & 91.43 & 99.82 & 81.12 ($\uparrow$ 2.19)                \\
                                      & 8                  & 56.07 & 60.89 & 68.21 & 80.18 & 84.29 & 85.18 & 96.07 & 75.84 ($\downarrow$ 4.75)                \\ \hline \hline
\end{tabular}
}
\label{tab:expofCF}
\end{table}

\begin{table}[]
\centering
\caption{The comparative experiments on Flight.}
\resizebox{0.50\textwidth}{!}{
\begin{tabular}{ccccccccc|c}
\hline
\hline
\multicolumn{1}{c}{\multirow{2}{*}{Algorithm}}    & & \multicolumn{7}{c|}{SNR}                              & \multirow{2}{*}{Average} \\ \cline{3-9}
\multicolumn{2}{c}{}                                         & -4    & 0     & 4     & 6     & 8     & 10    & 20    &                      \\ \hline
\multicolumn{1}{c|}{\multirow{4}{*}{Without TVD without MAF}} & Acc    & 52.14 & 58.04 & 71.25 & 71.79 & 89.11 & 88.57 & 98.75 & 75.66             \\
\multicolumn{1}{c|}{}                                & Pre    & 43.24 & 49.43 & 75.57 & 74.27 & 91.03 & 90.68 & 98.83 & 74.72             \\
\multicolumn{1}{c|}{}                                & F1     & 46.96 & 52.81 & 71.9  & 71.85 & 89.51 & 88.9  & 98.71 & 74.38             \\
\multicolumn{1}{c|}{}                                & Recall & 54.15 & 58.27 & 73.35 & 71.21 & 89.75 & 88.26 & 98.66 & 76.24             \\ \hline
\multicolumn{1}{c|}{\multirow{4}{*}{Without TVD With MAF}} & Acc    & 56.07 & 71.07 & 78.04 & 84.47 & 83.93 & 92.14 & 99.82 & 80.79             \\
\multicolumn{1}{c|}{}                                & Pre    & 47.87 & 72.92 & 86.24 & 88.15 & 84.89 & 93.26 & 99.82 & 81.88             \\
\multicolumn{1}{c|}{}                                & F1     & 49.05 & 71.8  & 76.62 & 84.67 & 84.42 & 92.07 & 99.82 & 79.78             \\
\multicolumn{1}{c|}{}                                & Recall & 58.87 & 72.64 & 79.19 & 85.31 & 84.16 & 92.87 & 99.81 & 81.84             \\ \hline
\multicolumn{1}{c|}{\multirow{4}{*}{With TVD Without MAF}} & Acc    & 68.57 & 75.71 & 82.86 & 85.89 & 91.78 & 94.82 & 98.57 & 85.46             \\
\multicolumn{1}{c|}{}                                & Pre    & 70.14 & 72.79 & 85.1  & 87.27 & 92.08 & 95.18 & 98.64 & 85.89         \\
\multicolumn{1}{c|}{}                                & F1     & 69.49 & 72.52 & 83.06 & 85.54 & 91.49 & 95    & 98.48 & 85.08             \\
\multicolumn{1}{c|}{}                                & Recall & 69.18 & 73.31 & 83.45 & 86.18 & 91.69 & 95.17 & 98.5  & 85.35             \\ \hline
\multicolumn{1}{c|}{\multirow{4}{*}{With TVD With MAF}} & Acc    & 71.25 & 74.46 & 83.21 & 87.32 & 93.39 & 96.61 & 100   & 86.61             \\
\multicolumn{1}{c|}{}                                & Pre    & 66.46 & 72.78 & 84.69 & 88.96 & 94.64 & 96.69 & 100   & 86.32             \\
\multicolumn{1}{c|}{}                                & F1     & 66.76 & 71.36 & 84.05 & 87.22 & 93.6  & 96.62 & 100   & 85.66             \\
\multicolumn{1}{c|}{}                                & Recall & 71.22 & 72.88 & 83.76 & 86.87 & 93.33 & 96.67 & 100   & 86.39                \\ \hline \hline
\end{tabular}
}
\label{tab:expofDF}
\end{table}

\section{CONCLUSION}
In addressing noise-induced challenges in mechanical system fault diagnosis, this study proposes the TDANet model, a novel advancement significantly enhancing diagnostic accuracy. During feature extraction, the STFT outperforms the FFT across two datasets. The inclusion of TVD and MAF modules further boosts diagnostic precision, especially in low SNR conditions. Compared to existing methods such as 1D CNN, CNN-ELM, and CNN-LSTM, the TDANet model demonstrates superior diagnostic accuracy. Future work will focus on validating the model's efficacy in more complex scenarios and exploring its practical deployment via mobile devices, aiming to enhance accessibility and effectiveness of fault diagnosis in diverse industrial contexts.


\bibliographystyle{Bibliography/IEEEtranTIE}
\bibliography{Bibliography/IEEEabrv,Bibliography/BIB_xx-TIE-xxxx}\ 

\begin{thebibliography}{10}
\providecommand{\url}[1]{#1}
\csname url@samestyle\endcsname
\providecommand{\newblock}{\relax}
\providecommand{\bibinfo}[2]{#2}
\providecommand{\BIBentrySTDinterwordspacing}{\spaceskip=0pt\relax}
\providecommand{\BIBentryALTinterwordstretchfactor}{4}
\providecommand{\BIBentryALTinterwordspacing}{\spaceskip=\fontdimen2\font plus
\BIBentryALTinterwordstretchfactor\fontdimen3\font minus \fontdimen4\font\relax}
\providecommand{\BIBforeignlanguage}[2]{{%
\expandafter\ifx\csname l@#1\endcsname\relax
\typeout{** WARNING: IEEEtran.bst: No hyphenation pattern has been}%
\typeout{** loaded for the language `#1'. Using the pattern for}%
\typeout{** the default language instead.}%
\else
\language=\csname l@#1\endcsname
\fi
#2}}
\providecommand{\BIBdecl}{\relax}
\BIBdecl

\bibitem{yang2019polynomial}
B.~Yang, Y.~Lei, F.~Jia, N.~Li, and Z.~Du, ``A polynomial kernel induced distance metric to improve deep transfer learning for fault diagnosis of machines,'' \emph{IEEE Transactions on Industrial Electronics}, vol.~67, no.~11, pp. 9747--9757, 2019.

\bibitem{he2021modified}
Z.~He, H.~Shao, Z.~Ding, H.~Jiang, and J.~Cheng, ``Modified deep autoencoder driven by multisource parameters for fault transfer prognosis of aeroengine,'' \emph{IEEE Transactions on Industrial Electronics}, vol.~69, no.~1, pp. 845--855, 2021.

\bibitem{chen2023novel}
J.~Chen, W.~Hu, G.~Zhang, J.~Hu, Q.~Huang, Z.~Chen, and F.~Blaabjerg, ``A novel knowledge sharing method for rolling bearing fault detection against impact of different signal sampling frequencies,'' \emph{IEEE Transactions on Instrumentation and Measurement}, 2023.

\bibitem{chen2021gaussian}
J.~Chen, W.~Hu, D.~Cao, M.~Zhang, Q.~Huang, Z.~Chen, and F.~Blaabjerg, ``Gaussian process kernel transfer enabled method for electric machines intelligent faults detection with limited samples,'' \emph{IEEE Transactions on Energy Conversion}, vol.~36, no.~4, pp. 3481--3490, 2021.

\bibitem{jalan2009model}
A.~K. Jalan and A.~Mohanty, ``Model based fault diagnosis of a rotor--bearing system for misalignment and unbalance under steady-state condition,'' \emph{Journal of sound and vibration}, vol. 327, no. 3-5, pp. 604--622, 2009.

\bibitem{dybala2014rolling}
J.~Dyba{\l}a and R.~Zimroz, ``Rolling bearing diagnosing method based on empirical mode decomposition of machine vibration signal,'' \emph{Applied Acoustics}, vol.~77, pp. 195--203, 2014.

\bibitem{zhang2020deep}
S.~Zhang, S.~Zhang, B.~Wang, and T.~G. Habetler, ``Deep learning algorithms for bearing fault diagnostics-a comprehensive review,'' \emph{IEEE Access}, vol.~8, pp. 29\,857--29\,881, 2020.

\bibitem{zarei2012induction}
J.~Zarei, ``Induction motors bearing fault detection using pattern recognition techniques,'' \emph{Expert systems with Applications}, vol.~39, no.~1, pp. 68--73, 2012.

\bibitem{qian2016application}
S.~Qian, X.~Yang, J.~Huang, and H.~Zhang, ``Application of new training method combined with feedforward artificial neural network for rolling bearing fault diagnosis,'' in \emph{2016 23rd International Conference on Mechatronics and Machine Vision in Practice (M2VIP)}, pp. 1--6.\hskip 1em plus 0.5em minus 0.4em\relax IEEE, 2016.

\bibitem{6030215}
M.~Xia, F.~Kong, and F.~Hu, ``An approach for bearing fault diagnosis based on pca and multiple classifier fusion,'' in \emph{2011 6th IEEE Joint International Information Technology and Artificial Intelligence Conference}, vol.~1, \href{http://dx.doi.org/10.1109/ITAIC.2011.6030215}{DOI 10.1109/ITAIC.2011.6030215}, pp. 321--325, 2011.

\bibitem{xue2017hybrid}
X.~Xue and J.~Zhou, ``A hybrid fault diagnosis approach based on mixed-domain state features for rotating machinery,'' \emph{ISA transactions}, vol.~66, pp. 284--295, 2017.

\bibitem{miao2021sparse}
M.~Miao, Y.~Sun, and J.~Yu, ``Sparse representation convolutional autoencoder for feature learning of vibration signals and its applications in machinery fault diagnosis,'' \emph{IEEE Transactions on Industrial Electronics}, vol.~69, no.~12, pp. 13\,565--13\,575, 2021.

\bibitem{janssens2016convolutional}
O.~Janssens, V.~Slavkovikj, B.~Vervisch, K.~Stockman, M.~Loccufier, S.~Verstockt, R.~Van~de Walle, and S.~Van~Hoecke, ``Convolutional neural network based fault detection for rotating machinery,'' \emph{Journal of Sound and Vibration}, vol. 377, pp. 331--345, 2016.

\bibitem{xu2021hybrid}
Y.~Xu, Z.~Li, S.~Wang, W.~Li, T.~Sarkodie-Gyan, and S.~Feng, ``A hybrid deep-learning model for fault diagnosis of rolling bearings,'' \emph{Measurement}, vol. 169, p. 108502, 2021.

\bibitem{mao2021new}
W.~Mao, W.~Feng, Y.~Liu, D.~Zhang, and X.~Liang, ``A new deep auto-encoder method with fusing discriminant information for bearing fault diagnosis,'' \emph{Mechanical Systems and Signal Processing}, vol. 150, p. 107233, 2021.

\bibitem{ma2019ensemble}
S.~Ma and F.~Chu, ``Ensemble deep learning-based fault diagnosis of rotor bearing systems,'' \emph{Computers in industry}, vol. 105, pp. 143--152, 2019.

\bibitem{chen2021bearing}
X.~Chen, B.~Zhang, and D.~Gao, ``Bearing fault diagnosis base on multi-scale cnn and lstm model,'' \emph{Journal of Intelligent Manufacturing}, vol.~32, no.~4, pp. 971--987, 2021.

\bibitem{zhang2019deep}
W.~Zhang, X.~Li, and Q.~Ding, ``Deep residual learning-based fault diagnosis method for rotating machinery,'' \emph{ISA transactions}, vol.~95, pp. 295--305, 2019.

\bibitem{liang2022intelligent}
P.~Liang, W.~Wang, X.~Yuan, S.~Liu, L.~Zhang, and Y.~Cheng, ``Intelligent fault diagnosis of rolling bearing based on wavelet transform and improved resnet under noisy labels and environment,'' \emph{Engineering Applications of Artificial Intelligence}, vol. 115, p. 105269, 2022.

\bibitem{guo2023rolling}
Y.~Guo, J.~Mao, and M.~Zhao, ``Rolling bearing fault diagnosis method based on attention cnn and bilstm network,'' \emph{Neural processing letters}, vol.~55, no.~3, pp. 3377--3410, 2023.

\bibitem{zhao2021new}
J.~Zhao, S.~Yang, Q.~Li, Y.~Liu, X.~Gu, and W.~Liu, ``A new bearing fault diagnosis method based on signal-to-image mapping and convolutional neural network,'' \emph{Measurement}, vol. 176, p. 109088, 2021.

\bibitem{sandsten2016time}
M.~Sandsten, ``Time-frequency analysis of time-varying signals and non-stationary processes,'' \emph{Lund University}, 2016.

\bibitem{hendriks2022towards}
J.~Hendriks, P.~Dumond, and D.~Knox, ``Towards better benchmarking using the cwru bearing fault dataset,'' \emph{Mechanical Systems and Signal Processing}, vol. 169, p. 108732, 2022.

\bibitem{li2023lightweight}
Z.~Li, Y.~Zhang, J.~Ai, Y.~Zhao, Y.~Yu, and Y.~Dong, ``A lightweight and explainable data-driven scheme for fault detection of aerospace sensors,'' \emph{IEEE Transactions on Aerospace and Electronic Systems}, 2023.

\bibitem{liu2021machinery}
H.~Liu, R.~Ma, D.~Li, L.~Yan, and Z.~Ma, ``Machinery fault diagnosis based on deep learning for time series analysis and knowledge graphs,'' \emph{journal of signal processing systems}, vol.~93, pp. 1433--1455, 2021.

\end{thebibliography}
	
\begin{IEEEbiography}[{\includegraphics[width=1in,height=1.25in,clip,keepaspectratio]{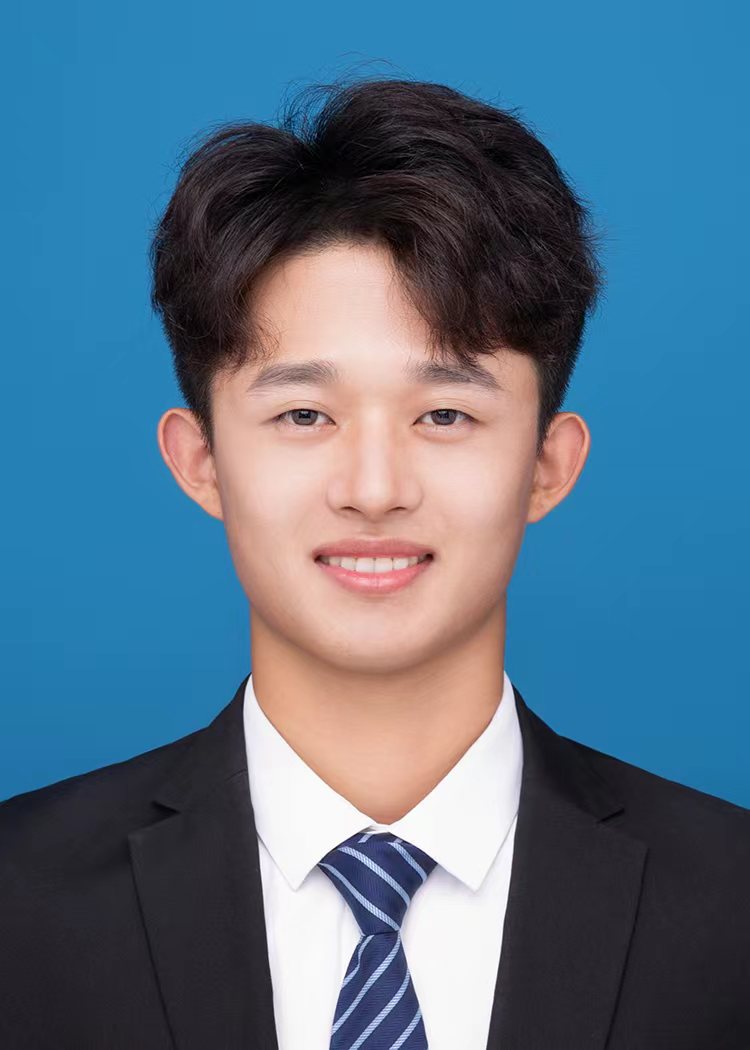}}]
{Zhongzhi Li} was born in Shandong, China, in 2000. He is presently working towards his M.S. degree in Department of Aeronautics and Astronautics at Fudan University, Shanghai, China. He has published more than 10 papers in journals and international conference proceedings. His current research interests include computer vision, intelligent fault detection and deep learning.
\end{IEEEbiography}

\begin{IEEEbiography}[{\includegraphics[width=1in,height=1.25in,clip,keepaspectratio]{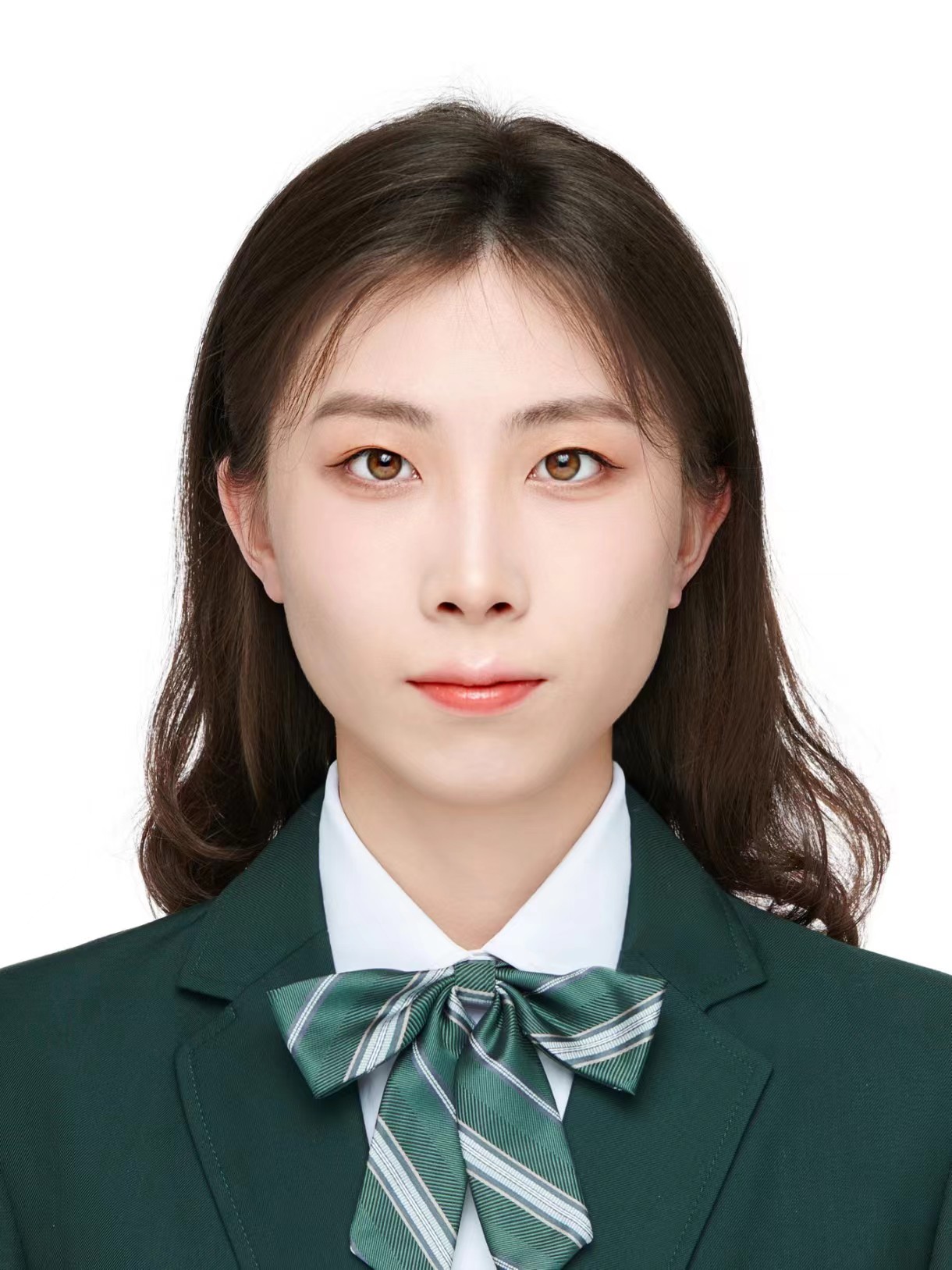}}]
{Rong Fan} was born in Henan, China, in 2003. She is presently a senior undergraduate at Shanghai University and is about to pursue her M.S. degree in Electronic Information in Department of Aeronautics and Astronautics at Fudan University, Shanghai, China. Her current research interests include data analysis and deep learning.
\end{IEEEbiography}

\begin{IEEEbiography}[{\includegraphics[width=1in,height=1.25in,clip,keepaspectratio]{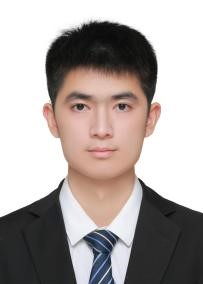}}]
{Jingqi Tu} was born in Jiangxi, China, in 2001. He is presently working towards his Ph.D degree in Department of Aeronautics and Astronautics at Fudan University, Shanghai, China.
His current research interests include data fusion and deep learning.
\end{IEEEbiography}

\begin{IEEEbiography}[{\includegraphics[width=1in,height=1.25in,clip,keepaspectratio]{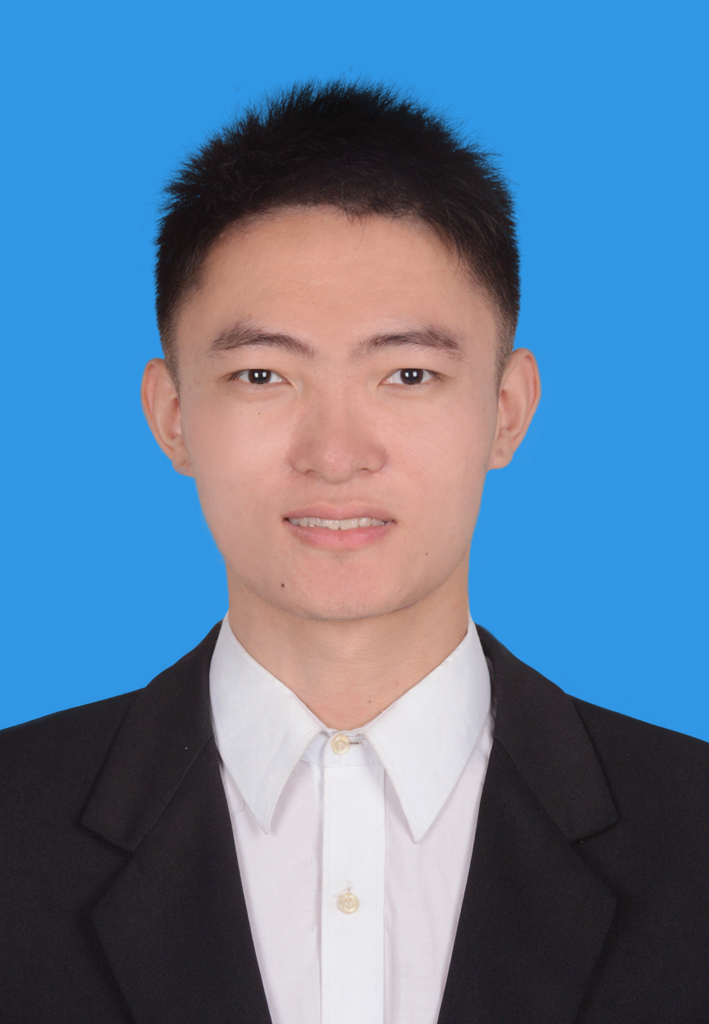}}]
{\bf Jinyi Ma} was born in Liaoning, China, in 1999. He is presently working towards his Ph.D. degree in Department of Aeronautics and Astronautics at Fudan University, Shanghai, China.

He has published more than 10 papers in journals and international conference proceedings. His current research interests include aircraft dynamics modeling, fault-tolerant control, and deep learning.
\end{IEEEbiography}

\begin{IEEEbiography}[{\includegraphics[width=1in,height=1.25in,clip,keepaspectratio]{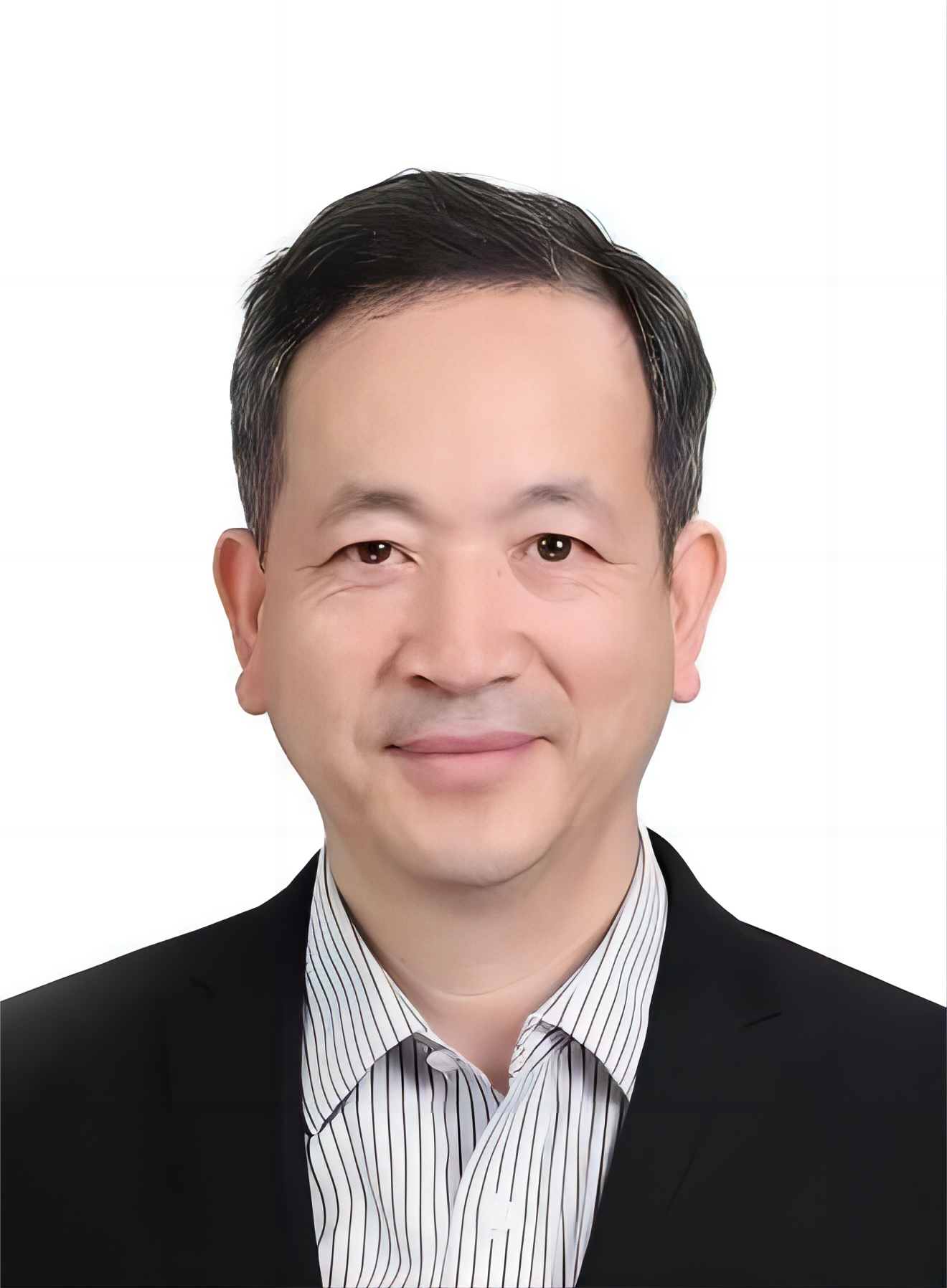}}]
{Jianliang Ai} received the B.E., M.S., and Ph.D. degrees in 1986, 1989, and 1997,
respectively, in flight dynamics and control, from the Department of Aerospace Design, Northwestern Polytechnical University, Xian, China.
He is a Professor with the Department of Aeronautics and Astronautics, Fudan University, Shanghai China. His research work is mainly on aircraft preliminary designs, with a specific focus on dynamics modeling and flight control laws design of the aircraft. He
is also heading the research work of exploring the capability of artificial intelligence in the
optimization stage of the aircraft design, dynamics identification, and intelligent control
laws applications. He has authored more than 50 highly cited journal papers.
\end{IEEEbiography}

\begin{IEEEbiography}[{\includegraphics[width=1in,height=1.25in,clip,keepaspectratio]{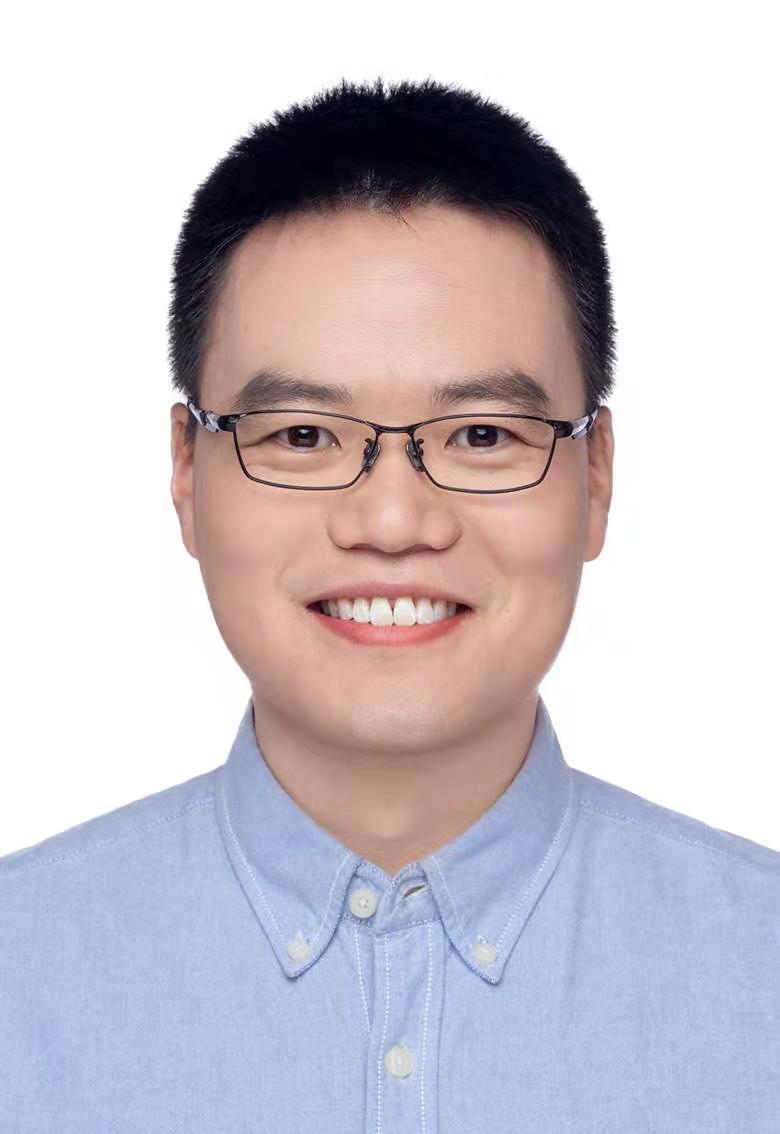}}]
{Yiqun Dong} was born in Jiangsu, China,
in 1990. He received the B.E. and Ph.D. degrees
in aerospace engineering from Fudan University,
Shanghai, China, in 2010 and 2016, respectively.
From 2014 to 2016, he was an Exchange
Ph.D. Student with the Department of Mechanical,
Industrial, and Aerospace Engineering, Concordia
University, Montreal, QC, Canada. From 2016 to
2017, he was a Postdoctoral Research Fellow with
the Department of Electrical and Electronic Engineering, Nanyang Technological University, Singapore. From 2017 to 2019,
he was a Postdoctoral Research Fellow with the Department of Mechanical
Engineering, University of Michigan, Ann Arbor, MI, USA. Since 2019,
he has been an Associate Professor with the Department of Aeronautics and
Astronautics, Fudan University. He has published more than 60 highly-cited
journal/conference papers. His research interests include intelligent fault detection and classification methods.
\end{IEEEbiography}

\end{document}